\documentclass[10pt,twocolumn,letterpaper]{article}

\usepackage[pagenumbers]{cvpr} 
\usepackage{multirow}
\usepackage{siunitx}
\usepackage{amsmath}
\usepackage{arydshln}
\usepackage{comment}
\usepackage{pifont}
\newcommand{\cmark}{\ding{51}}
\newcommand{\xmark}{\ding{55}}
\usepackage{booktabs,amsfonts}
\newcommand{\std}[1]{{\footnotesize$_{\pm#1}$}}

\usepackage{natbib}
\usepackage{xcolor} 
\usepackage{float} 
\usepackage{svg}

\definecolor{cvprblue}{rgb}{0.21,0.49,0.74}
\usepackage[pagebackref,breaklinks,colorlinks,allcolors=cvprblue]{hyperref}
\def\paperID{} 
\def\confName{CVPR}
\def\confYear{2026}

\title{From Words to Wavelengths: VLMs for \\Few-Shot Multispectral Object Detection}

\author{
Manuel Nkegoum${}^{1,3}$, Minh-Tan Pham${}^1$,  Élisa Fromont${}^2$, Bruno Avignon${}^3$ and Sébastien Lefèvre${}^{1,4}$\\ \\
${}^1$Univ Bretagne Sud, IRISA, UMR 6074, Vannes, France\\
${}^2$Univ Rennes, IRISA, UMR 6074, Rennes, France\\
${}^3$ATERMES, Montigny-le-Bretonneux, France \\
${}^4$UiT The Arctic University of Norway, Tromsø, Norway \\
\tt\small \{manuel.nkegoum-nzouakeu,minh-tan.pham,elisa.fromont,sebastien.lefevre\}@irisa.fr, \\
\tt\small bavignon@atermes.fr
}

\begin{document}
\maketitle
\begin{abstract}
Multispectral object detection is essential for safety-critical systems but remains challenging due to the scarcity of annotated RGB-IR training pairs. While existing methods invest in increasingly sophisticated cross-modal fusion modules, the 
few-shot multispectral setting remains largely underexplored. We investigate Vision-Language Models (VLMs) for few-shot multispectral object detection (FSMOD) and argue that the shared textual supervision implicitly aligns RGB and IR representations under unified category semantics, reducing the need for complex fusion design. We adapt two state-of-the-art VLM-based detectors to multispectral inputs and show that (i) simple fusion mechanisms can match or outperform complex cross-modal fusion strategies when built upon strong VLM 
representations, and (ii) VLMs trained solely on RGB data can generalize effectively to unseen thermal modalities without backbone adaptation. Experiments on FLIR, M3FD, and DroneVehicle show improvements of up to 30~mAP@50 over existing methods. All codes and original data splits can be found at \href{https://github.com/Manuelnkegoum-8/MSVLM.git}{https://github.com/Manuelnkegoum-8/MSVLM}
\end{abstract}
\section{Introduction}
\label{sec:intro}

\begin{figure}[tb]
    \centering
    \includegraphics[width=0.45\textwidth]{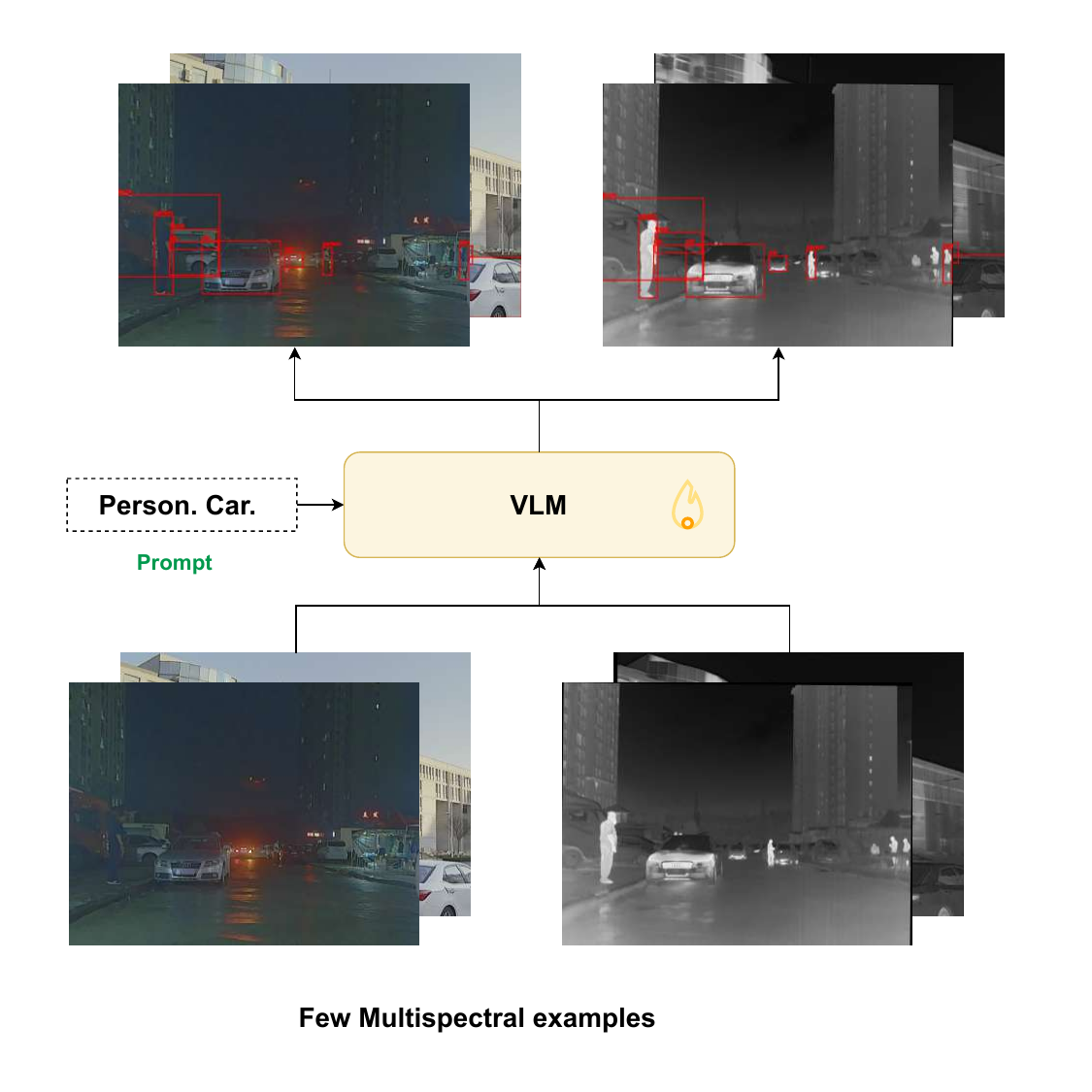}
    \caption{Conventional multispectral detectors require many annotated, aligned RGB-IR pairs for training. Our approach leverages the semantic priors of vision-language models (VLMs) to bridge this data gap. Given only a few multispectral examples and a text prompt (e.g., ``person, car''), the adapted VLM can robustly detect objects across spectral domains by jointly reasoning over visual (RGB) and thermal (IR) cues.}
    \label{fig:placeholder}
\end{figure}

Reliable perception under challenging conditions such as nighttime, fog, or occlusion is critical for autonomous driving, surveillance, and robotics. Multispectral object detection addresses this challenge by leveraging complementary information from visible (RGB) and infrared (IR) modalities, enabling robust detection even when one modality degrades. While the term 
\emph{multispectral} broadly encompasses various sensor combinations, this 
work focuses on the RGB-thermal setting, which is the most prevalent in 
safety-critical applications and supported by established benchmarks.
Yet large-scale aligned RGB-IR annotations remain scarce, since acquiring 
accurately registered image pairs with bounding boxes is expensive and 
time-consuming. Vision-Language Models (VLMs) have recently demonstrated strong few-shot and open-vocabulary detection capabilities in the RGB domain~\cite{Madan2024Revisiting,liu2024grounding}, acting as powerful backbones that generalize to novel object classes with minimal supervision. Inspired by this success, we investigate whether the semantic grounding learned from RGB image-text pairs can transfer to multispectral inputs and support effective few-shot detection across spectral modalities.

In this work, we investigate VLM-based detectors for few-shot multispectral 
object detection and question whether the increasingly sophisticated cross-modal 
fusion strategies proposed in the literature are truly needed. We extend two 
representative VLM frameworks, Grounding DINO~\cite{liu2024grounding} and 
YOLO-World~\cite{cheng2024yolo}, to jointly process RGB and IR inputs, and 
conduct a systematic study of fusion strategies of increasing complexity, from parameter-free summation to learned attention-based fusion. Our central finding is that the shared textual supervision implicitly aligns RGB and IR 
representations under unified category semantics such that simple fusion consistently matches or outperforms learned fusion strategies across all data regimes, from 5-shot to full supervision. \Cref{sec:related} reviews related work on object detection, multispectral detection, few-shot learning, and VLMs for detection. \Cref{sec:method} details our multispectral extension and the studied fusion paradigms. \Cref{sec:experiments} presents our experimental evaluation. \Cref{sec: conclusion} concludes the paper.

\section{Related Work}
\label{sec:related}

\subsection{Object Detection}

Object detection has witnessed remarkable progress driven by deep learning architectures. Two-stage detectors like Faster R-CNN~\cite{ren2015faster} pioneered region proposal mechanisms followed by classification and regression. Subsequent one-stage approaches including SSD~\cite{liu2016ssd} and  YOLO~\cite{redmon2016yolo} achieved improved efficiency through unified prediction frameworks. More recently, transformer-based detectors such as DINO~\cite{zhang2022dino} and RT-DETR~\cite{zhao2024detrs} have demonstrated superior performance by leveraging self-attention mechanisms and dynamic object queries. Collaborative frameworks like Co-DETR~\cite{zong2023detrs} further enhanced detection performance through multi-view supervision and auxiliary training signals. These architectural advances have established a solid foundation for high-performance object detection, providing the building blocks for multispectral extensions.

\subsection{Multispectral Object Detection}

Multispectral object detection exploits complementary information from different imaging modalities to enhance robustness under challenging conditions. Early fusion approaches simply concatenated features from different spectra, while more sophisticated methods developed specialized fusion strategies. GAFF~\cite{zhang2021guided} introduced Guided Attentive Feature Fusion with inter- and intra-modality attention mechanisms to dynamically combine multispectral features. DAMSDet~\cite{guo2025damsdet} introduces a dynamic multispectral detection transformer with adaptive fusion. Most recently, GM-DETR~\cite{xiao2024gm} enhanced global context modeling across spectral channels, improving detection consistency across varying illumination conditions. These methods collectively demonstrate that sophisticated cross-modal interaction strategies are essential for high-performance multispectral detection.

\subsection{Few-Shot Object Detection}

Few-shot object detection (FSOD) aims to recognize novel object categories with limited labeled examples. Early FSOD methods \cite{yan2019meta} employed meta-learning strategies to transfer knowledge from base to novel classes. To date, only two methods have specifically addressed few-shot learning in multispectral domains (FSMOD): FSMODNet~\cite{nkegoum2025fsmodnet}, which 
tackles FSMOD through cross-modal feature fusion, and Cross-Modality 
Interaction~\cite{huang2024cross}, which relies on semantic prototype alignment. Our work diverges by investigating VLM-based approaches, which offer stronger generalization through semantic grounding.

\subsection{Vision-Language Models for Detection}

VLMs have revolutionized open-vocabulary object detection by aligning visual representations with semantic embeddings. GLIP~\cite{li2022grounded} unified object detection and phrase grounding through language-image pretraining, enabling zero-shot transfer to novel categories. YOLO-World~\cite{cheng2024yolo} extended the YOLO architecture with language-conditioned heads for prompt-based adaptation. GroundingDINO~\cite{liu2024grounding} achieved state-of-the-art performance by grounding dynamic object queries in natural language embeddings. These models demonstrate that semantic knowledge from large-scale pretraining can effectively transfer to detection tasks with minimal fine-tuning, motivating our exploration of VLM capabilities in multispectral domains.

\begin{figure*}[tb]
\centering
\scalebox{1}{
\includegraphics[width=\textwidth]{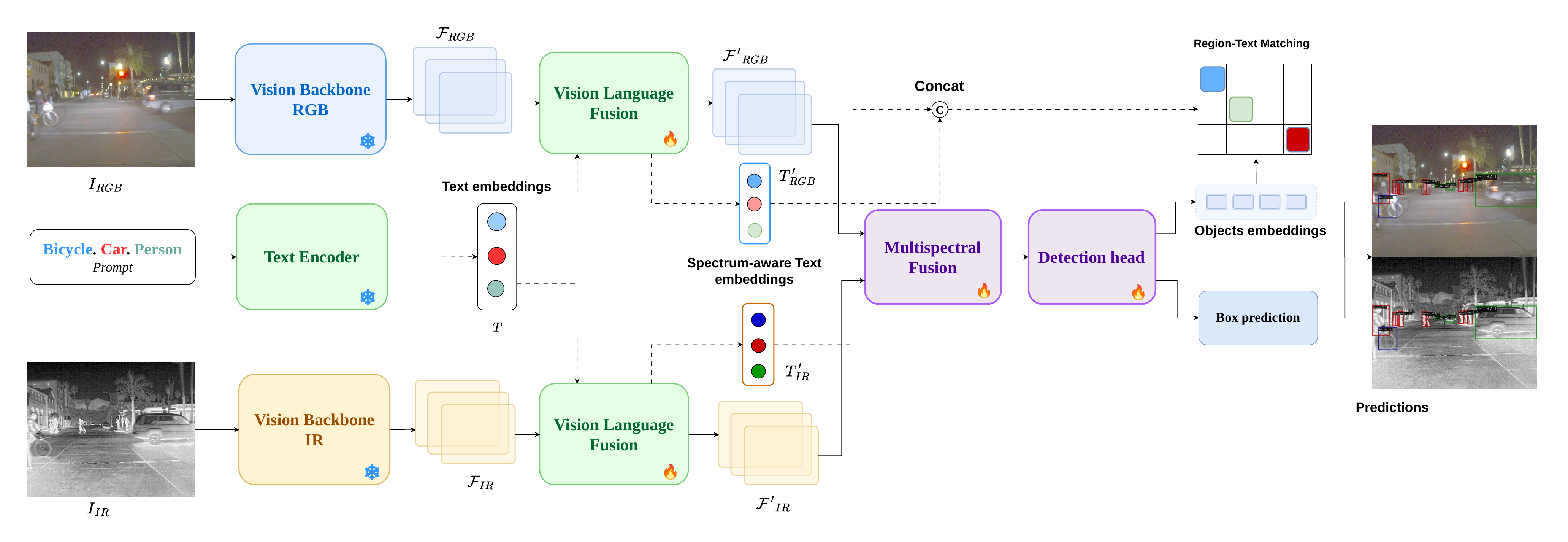}
}
\caption{Overall architecture of MS-X instantiated using either 
\textbf{Grounding DINO (MS-GDINO)} or \textbf{YOLO-World (MS-YOLOW)}. The model processes RGB and IR inputs through separate backbones and encoders, with fusion occurring at the query selection and decoder stages. Text embeddings provide semantic guidance throughout the pipeline.}
\label{fig:architecture}
\end{figure*}

\section{Methodology}
\label{sec:method}
We investigate the integration of VLMs into the few-shot multispectral object detection setting by extending language-guided detection frameworks to jointly process RGB and infrared inputs. To this end, we examine two representative VLM-based paradigms, a transformer-based detector (MS-GDINO) and a convolutional detector (MS-YOLOW), within a unified multispectral formulation. Despite their architectural differences, both models follow a common principle illustrated in \cref{fig:architecture}: textual prompts provide semantic supervision while visual representations are learned from aligned RGB-IR inputs. This design enables the study of how semantic priors acquired through large-scale RGB-text pretraining transfer to the multispectral domain under limited supervision. Formally, let $(I_{\text{RGB}}, I_{\text{IR}})$ denote an aligned multispectral image pair and $\mathcal{P}$ a textual prompt describing the target classes (e.g., ``person'', ``car'', \ldots). The following sections detail how visual and textual features are integrated and evaluated in the few-shot multispectral regime.

\subsection{Cross-Spectral Semantic Transfer}
\label{sec:cross_spectral_transfer}

\begin{figure*}[tb]
    \centering
    \includegraphics[width=0.7\linewidth]{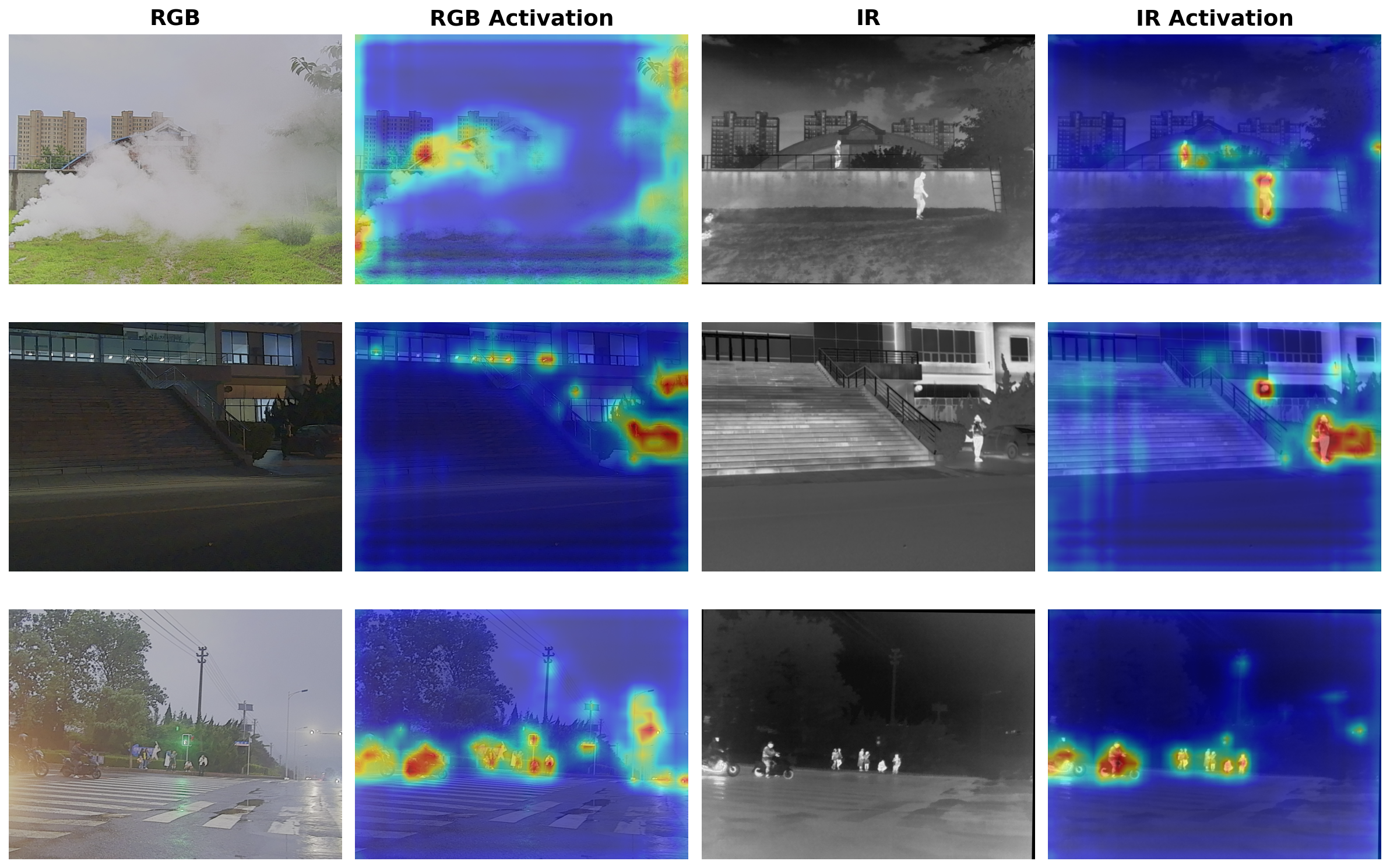}
    \caption{
Semantic activation maps from a frozen YOLO-World-M backbone on M3FD image pairs using the medium-scale features. The prompts used come fromcthe M3FD class set: \textbf{person, car, bus, truck, etc}. Maps are generated from the text cross-attention layers (\texttt{MaxSigmoidAttn}) of YOLO-World-M.
Despite the modality gap, both RGB and thermal inputs activate consistent object regions, suggesting that pretrained VLM features retain transferable semantic structure across spectra.
}
    \label{fig:heatmaps}
\end{figure*}

A key question in adapting vision-language models to multispectral object detection is whether models pretrained on RGB data can directly operate on infrared inputs without backbone finetuning or architectural modification. Although RGB and infrared modalities differ in low-level appearance statistics, they share consistent high-level semantic structure, such as object shape and spatial layout. To examine this, we visualize intermediate semantic activations  of a pretrained YOLO-World model on aligned RGB-IR image pairs from M3FD~\cite{liu2023m3fd}, using a frozen RGB-pretrained backbone. As shown in \cref{fig:heatmaps}, despite the absence of infrared-specific training the thermal modality clearly activates the pedestrian region behind the smoke, while the corresponding region is substantially less pronounced in the RGB modality. This behavior suggests that pretrained vision-language representations retain sufficient semantic structure to support object localization across spectral domains. Based on this observation, we do not introduce any infrared-specific adaptation or backbone modification, and instead directly apply the same pretrained vision-language encoder to both modalities.

\subsection{Dual-Modality Encoding}
\label{sec:dual_modality_encoding}

We extract multi-scale feature representations from each modality using modality-specific backbones:
\begin{align}
\mathcal{F}_{m} = \mathrm{Backbone}_{m}(I_{m}), \quad m \in \{\mathrm{RGB}, \mathrm{IR}\},
\end{align}
where $\mathcal{F}_{m} = \{F_{m}^{1}, \dots, F_{m}^{L}\}$ and $F_{m}^{l} \in 
\mathbb{R}^{C_l \times H_l \times W_l}$ denotes the feature map at level $l$. 
The prompt $\mathcal{P}$ is encoded by a pretrained language model, BERT~\cite{devlin2019bert} 
for MS-GDINO and the CLIP text encoder~\cite{radford2021learning} for MS-YOLOW:
\begin{align}
T = \mathrm{TextEncoder}(\mathcal{P}) \in \mathbb{R}^{N_t \times d},
\end{align}
where $N_t$ is the number of tokens and $d$ the embedding dimension. Each modality 
stream is processed by a vision-language encoder specifically the feature enhancer 
from Grounding DINO~\cite{liu2024grounding} for MS-GDINO, and the RepConv-based 
encoder from YOLO-World~\cite{cheng2024yoloworld} for MS-YOLOW:
\begin{align}
\mathcal{F}'_{m}, T'_{m} = \mathrm{Encoder}_{m}(\mathcal{F}_{m}, T), 
\quad m \in \{\mathrm{RGB}, \mathrm{IR}\},
\end{align}
producing refined visual features $\mathcal{F}'_{m}$ enriched with language 
guidance and modality-conditioned text features $T'_{m}$.

\subsection{Multispectral Fusion Module}
\label{sec:fusion_module}

Given the modality-specific enhanced features $\mathcal{F}'_{\mathrm{RGB}}$ and $\mathcal{F}'_{\mathrm{IR}}$ from the dual vision-language encoders, we investigate different fusion paradigms of increasing complexity to study the trade-off between fusion expressiveness and data efficiency in few-shot multispectral detection.
\paragraph{Parameter-Free Fusion (sum).}
At first, we consider element-wise summation, which introduces no 
learnable parameters:
\begin{align}
\mathbf{F}_{\mathrm{fused}} = \mathcal{F}'_{\mathrm{RGB}} + \mathcal{F}'_{\mathrm{IR}}.
\end{align}
This operation directly aggregates both modality features at each spatial location without additional transformations.

\paragraph{Lightweight Learnable Fusion (conv).}
As a second paradigm, we consider channel concatenation followed by a $1\times1$ 
convolution to restore the original channel dimensionality:
\begin{align}
\mathbf{F}_{\mathrm{fused}} = \mathrm{Conv}_{1\times1}\!\left(
    \mathrm{Concat}\!\left(\mathcal{F}'_{\mathrm{RGB}},\, \mathcal{F}'_{\mathrm{IR}}\right)
\right).
\end{align}
This introduces a minimal number of learnable parameters, allowing the model to 
learn channel-wise modality weights while remaining lightweight enough to train 
under limited supervision.

\paragraph{Differential Cross-Attention (cross-att).}
As a third paradigm, we explore a bidirectional cross-attention fusion module inspired by the differential attention mechanism~\cite{ye2025differential}. Unlike standard cross-attention, which aggregates cross-modal interactions through a single attention map, differential attention introduces a contrastive formulation between two parallel attention maps to emphasize differences in learned responses.
We adopt this mechanism for multispectral fusion by applying it to RGB and IR feature interactions in a bidirectional manner. This constitutes our most expressive fusion design, at the cost of significantly more learnable parameters. Full implementation details are provided in the supplementary material (\cref{sec:fusion_para}). We additionally compare against the text-guided consensus-discrepancy 
fusion of~\cite{wu2026bridging} (cons-disc) as a strong external baseline, 
which explicitly decomposes RGB-IR interactions into consensus and discrepancy 
support patterns via learned recalibration.

\subsection{Prediction Head}
\label{sec:prediction_head}

The prediction head operates on fused visual features $\mathbf{F}_{\mathrm{fused}}$ and modality-conditioned text embeddings $T'_{\mathrm{RGB}}, T'_{\mathrm{IR}}$.

\paragraph{MS-GDINO.}
MS-GDINO refines object queries $\mathbf{Q}_{\mathrm{out}}$ through a transformer decoder that attends to $\mathbf{F}_{\mathrm{fused}}$. Bounding boxes are predicted as:
\begin{align}
\hat{\mathbf{b}} = \mathrm{MLP}_{\mathrm{box}}(\mathbf{Q}_{\mathrm{out}}),
\end{align}
while class predictions are obtained by computing similarity between query features and both modality-conditioned text embeddings:
\begin{align}
\hat{\mathbf{p}} = \max_{m \in \{\mathrm{RGB},\mathrm{IR}\}} \mathbf{Q}_{\mathrm{out}} (T'_m)^\top.
\end{align}

The maximum operation selects the strongest modality-aligned response for each prediction, allowing the model to remain robust to modality-specific variations without introducing additional classification heads.

\paragraph{MS-YOLOW.}
MS-YOLOW performs dense detection directly on multi-scale fused features without object queries. At each feature level $l$, bounding box and class predictions are computed as:
\begin{align}
\hat{\mathbf{b}}^l &= \mathrm{Conv}_{\mathrm{box}}(\mathbf{F}_{\mathrm{fused}}^l), \\
\hat{\mathbf{p}}^l &= \max_{m \in \{\mathrm{RGB},\mathrm{IR}\}} \mathbf{F}_{\mathrm{fused}}^l (T'_m)^\top.
\end{align}

As in MS-GDINO, the maximum over modalities is used to select the most responsive text alignment, ensuring consistent predictions across RGB and infrared inputs.


\section{Experiments}
\label{sec:experiments}

We present a comprehensive experimental evaluation of our framework across 
multiple datasets, data regimes, and model configurations. Our evaluation is 
structured around three axes: (i) few-shot performance under extreme annotation 
scarcity, where we compare against specialized multispectral detectors and 
unimodal VLM baselines; (ii) a systematic fusion study that empirically 
validates our central hypothesis that textual supervision subsumes the role 
of visual fusion design; and (iii) full supervision scalability, establishing 
upper-bound performance against state-of-the-art multispectral methods. 
Additional analyses examine the role of pseudo-labeling, the effect of 
RGB-IR misalignment, zero-shot transfer, and runtime efficiency.

\subsection{Datasets}
Our evaluation employs two challenging multispectral benchmarks that provide precisely aligned RGB-IR image pairs under diverse environmental conditions. \\
\textbf{FLIR}~\cite{flir2019} consists of synchronized RGB-thermal image pairs collected in automotive environments. We use the aligned subset from \cite{ZHANG_2020_ICIP}, containing 4,129 training and 1,013 test pairs. \\
\textbf{M3FD}~\cite{liu2023m3fd} provides 4,200 co-registered RGB-thermal pairs under diverse conditions, including day, night, overcast, and challenging scenes. Following \cite{guo2025damsdet}, we adopt its standard split for fair comparison. In addition, we also performed experiments on \textbf{DroneVehicle}~\cite{sun2020drone}, an aerial RGB-thermal dataset for vehicle detection in complex urban and highway scenarios, with large variations in scale and viewpoint. The results are given in the supplementary materials (see. \cref{tab:dronevehicle_fewshot_fusion}). 


\subsection{Few-Shot Learning Protocol}
To evaluate data-efficient learning, we follow the few-shot detection protocol of FSMODNet~\cite{nkegoum2025fsmodnet}, which provides standardized splits for multispectral datasets. For each dataset and category, we use predefined support sets with $k \in \{5,10\}$ annotated instances. Unless otherwise specified, experiments use the default fusion configuration introduced in Section~\ref{sec:fusion_module}. We additionally analyze the impact of fusion design through controlled comparisons with alternative mechanisms, including lightweight concatenation-based fusion and cross-attention-based fusion, as defined in the fusion study. We compare against two state-of-the-art multispectral detectors, DAMSDet~\cite{guo2025damsdet} and CAFF-DINO~\cite{helvig2024caff}. All baselines are initialized from COCO-pretrained weights and fine-tuned using their official implementations. This differs from the original FSMODNet setting, where models are trained from scratch, providing a stronger and more practical initialization regime across all methods.

\subsection{Evaluation Metrics}

Detection performance is quantified using established metrics appropriate for each experimental setting. For few-shot evaluation, we report mean Average Precision at IoU threshold 0.50 (mAP50), which provides a robust measure of detection accuracy in low-data regimes. Fully supervised experiments employ the comprehensive COCO evaluation suite: mAP, mAP50, and mAP75.

\subsection{Implementation Details}

All experiments are implemented using the MMDetection~\cite{mmdetection} and MMYolo~\cite{mmyolo2022} codebases built on PyTorch, and trained on NVIDIA A100 GPUs. Both visual backbones and text encoders are kept frozen during training to maintain the integrity of pre-trained representations and prevent overfitting in few-shot scenarios. During inference, we evaluate models at a single resolution of $640 \times 640$. We train our models for 50 epochs with an initial learning rate of $1 \times 10^{-5}$ for MS-GDINO and $5 \times 10^{-5}$ for MS-YOLOW. We use the \textbf{Tiny} variant for Grounding DINO and the \textbf{Medium} variant of YOLOW as bigger models were subject to severe overfitting in early experiments. Each experiment is repeated ten times with different random seeds, and the average score is reported to account for variance due to random sampling.

\textbf{Pseudo-Labeling.} 
Few-shot object detection suffers from \emph{incomplete supervision}, where many valid objects remain unlabeled because only a few instances per class are annotated. This issue is particularly severe in multispectral and densely populated datasets such as FLIR and M3FD. Training a VLM on such data implicitly teaches the model to interpret unlabeled objects as background, introducing \emph{false-negative supervision} that reduces recall and harms generalization. To alleviate this, we employ a \emph{pseudo-labeling} strategy to enrich the sparse annotations. For each image, let $\mathcal{G}$ denote its set of ground-truth boxes, and let the model generate candidate detections $\{(b_j, s_j, c_j)\}_j$ (box, score, class), retaining only those above an adaptive threshold:
\begin{align}
\tau = \max(\mu + \sigma, \tau_{\text{floor}}),
\end{align}
where \(\mu\) and \(\sigma\) denote the mean and standard deviation of detection scores \(\{s_j\}\) for the image. 
The lower bound of $\tau_{\text{floor}}$ prevents low-confidence predictions from contaminating the pseudo-label set when the model is uncertain. We found in ablations (\cref{sec:tau}) that the value of \(0.35\) works well in practice. After applying class-specific non-maximum suppression (NMS), a candidate \((b_j, s_j, c_j)\) is accepted as a pseudo-label only if it does not significantly overlap any ground-truth box of the same class:
\begin{align}
\max_{\substack{g \in \mathcal{G} \\ \text{class}(g) = c_j}} \mathrm{IoU}(b_j, g) < \delta,
\end{align}
where we set $\delta = 0.3$ empirically. Accepted pseudo-labels are merged with the original annotations and used as additional positives during training. This procedure allows the model to recover missing annotations, leveraging its zero-shot detection ability while keeping pseudo-label noise minimal.

\subsection{Results}
\label{sec:results}

\subsubsection{Few-Shot Multispectral Object Detection}
\label{sec:fsmod_results}

\paragraph{Results on FLIR.}
\cref{tab:flir_fewshot_perclass} summarizes performance on FLIR. Both multispectral 
variants consistently outperform their unimodal counterparts across all shot 
settings. In the 5-shot regime, MS-YOLOW achieves 70.69 mAP50, surpassing 
YOLOW-M~(RGB) and YOLOW-M~(IR) by +14.88 and +1.00 points respectively. 
MS-GDINO reaches 69.88 mAP50, improving over GDINO-T~(RGB) by +12.59 and 
GDINO-T~(IR) by +1.84 points. Crucially, both multispectral models substantially 
outperform the specialized multispectral detectors DAMSDet and CAFF-DINO, which 
achieve only 40.64 and 27.83 mAP50, highlighting the benefit of VLM semantic 
representations over task-specific architectures under scarce supervision.
\paragraph{Results on M3FD.}
\cref{tab:m3fd_fewshot_perclass} reports results on M3FD, which covers more 
diverse categories and challenging illumination conditions. The trends are 
consistent with FLIR. In the 5-shot setting, MS-GDINO achieves 66.33 mAP50, 
outperforming GDINO-T~(RGB) by +0.35 and GDINO-T~(IR) by +26.33 points. 
MS-YOLOW surpasses YOLOW-M~(RGB) by +2.73 and YOLOW-M~(IR) by +22.91 points. 
At 10-shot, MS-GDINO reaches 67.16 mAP50, the best overall result. The 
consistent gains across detectors, datasets, and conditions confirm that 
our approach generalizes well to heterogeneous spectral distributions.
\paragraph{Progressive improvements.}
Both frameworks exhibit consistent gains when scaling from 5 to 10 shots. On FLIR, MS-YOLOW improves from 70.69 to 71.15 mAP50 (+0.46) and MS-GDINO from 69.88 to 70.25 (+0.37). On M3FD, MS-YOLOW increases from 62.19 to 62.69 (+0.50) and MS-GDINO from 66.33 to 67.16 (+0.83). These moderate but consistent gains reflect two complementary effects: pseudo-labeling amplifies supervision by densifying annotations for underrepresented classes, while the pretrained VLM representations saturate quickly additional labels yield steady incremental improvements rather than large jumps, reflecting the models' robustness under 
low-data conditions.

\renewcommand{\arraystretch}{1.2}
\begin{table*}[tb]
\small
\centering

\setlength{\tabcolsep}{3.5pt} 

\scalebox{1.0}{\begin{tabular}{@{}llccccc@{}}
\toprule
Shots & Method & Spectrum & Bicycle & Car & Person & All \\
\midrule

\multirow{8}{*}{5-shot} 
 & DAMSDet   & RGB+IR & 36.48\std{8.23} & 43.12\std{7.49} & 42.32\std{6.27} & 40.64\std{3.74} \\
 & CAFF-DINO & RGB+IR & 18.00\std{7.74} & 40.30\std{6.65} & 25.20\std{4.50} & 27.83\std{2.04} \\
 \cdashline{2-7}

 & YOLOW-M   & RGB    & 43.54\std{3.44} & 68.40\std{1.11} & 55.50\std{1.07} & 55.81\std{1.15} \\
 & YOLOW-M   & IR     & 54.47\std{1.82} & 77.88\std{0.88} & \textbf{76.66}\std{0.57} & 69.69\std{0.55} \\
 & MS-YOLOW-M & RGB+IR & \textbf{58.00}\std{1.98} & \textbf{79.01}\std{1.79} & \underline{75.05}\std{0.56} & \textbf{70.69}\std{0.36} \\
 \cdashline{2-7}

 & GDINO-T   & RGB    & 44.72\std{1.39} & 69.71\std{1.14} & 58.89\std{0.70} & 57.29\std{0.84} \\
 & GDINO-T   & IR     & 52.50\std{1.20} & 76.34\std{1.61} & 74.54\std{0.79} & 68.04\std{1.11} \\
 & MS-GDINO-T & RGB+IR & \underline{57.63}\std{0.69} & \underline{78.37}\std{0.81} & 73.67\std{0.74} & \underline{69.88}\std{0.65} \\
\midrule

\multirow{8}{*}{10-shot} 
 & DAMSDet   & RGB+IR & 39.69\std{11.40} & 50.31\std{8.20} & 52.19\std{4.40} & 47.40\std{5.09} \\
 & CAFF-DINO & RGB+IR & 24.30\std{7.36}  & 40.50\std{4.69} & 31.90\std{4.71} & 32.23\std{1.78} \\
 \cdashline{2-7}

 & YOLOW-M   & RGB    & 43.14\std{2.16} & 69.18\std{1.16} & 56.25\std{1.89} & 56.18\std{1.08} \\
 & YOLOW-M   & IR     & 56.85\std{1.73} & 77.92\std{0.64} & \textbf{76.65}\std{0.99} & 69.81\std{0.77} \\
 & MS-YOLOW-M & RGB+IR & \textbf{59.37}\std{2.92} & \underline{79.58}\std{1.68} & 74.60\std{0.81} & \textbf{71.15}\std{0.59} \\
 \cdashline{2-7}

 & GDINO-T   & RGB    & 44.84\std{1.18} & 70.59\std{1.20} & 59.37\std{1.08} & 58.27\std{0.88} \\
 & GDINO-T   & IR     & 51.98\std{1.86} & 76.79\std{0.55} & 74.69\std{0.85} & 67.82\std{0.80} \\
 & MS-GDINO-T & RGB+IR & \underline{58.66}\std{3.25} & \textbf{79.88}\std{1.14} & \underline{74.97}\std{1.26} & \underline{70.25}\std{1.36} \\
\bottomrule
\end{tabular}
}
\caption{Few-shot detection on FLIR. mAP50 reported. Standard deviations over 10 runs shown as subscripts. Bold and underline indicate best and second-best per shot setting.}
\label{tab:flir_fewshot_perclass}
\end{table*}
\renewcommand{\arraystretch}{1.0}

\renewcommand{\arraystretch}{1.2}

\begin{table*}[tb]
\small
\centering
\setlength{\tabcolsep}{3.5pt}
\scalebox{1.0}{
\begin{tabular}{@{}llcccccc@{}}
\toprule
Shots & Method & Spectrum & Person & Car & Bus & Moto & All \\
\midrule
\multirow{8}{*}{5-shot}
 & DAMSDet        & RGB+IR & 13.26\std{4.12} & 36.12\std{6.65} & 30.36\std{5.33} & 27.39\std{10.12} & 27.86\std{3.19} \\
 & CAFF-DINO      & RGB+IR & 8.21\std{3.51}  & 31.92\std{5.84} & 35.56\std{8.12} & 18.56\std{10.37} & 22.48\std{2.90} \\
\cdashline{2-8}
 & YOLOW-M        & RGB    & 44.10\std{0.47} & 84.02\std{0.52} & \underline{70.59}\std{2.13} & 63.58\std{2.48}  & \underline{65.98}\std{1.00} \\
 & YOLOW-M        & IR     & 49.33\std{1.92} & 67.88\std{0.91} & 48.00\std{6.95} & 27.40\std{5.52}  & 39.28\std{1.56} \\
 & \bf MS-YOLOW-M & RGB+IR & 52.36\std{1.31} & 84.81\std{1.17} & \underline{70.59}\std{4.56} & 58.01\std{3.33} & 62.19\std{0.83} \\
\cdashline{2-8}
 & GDINO-T        & RGB    & 47.82\std{0.67} & \textbf{86.97}\std{0.73} & 67.82\std{1.51} & \underline{69.74}\std{1.31} & 65.98\std{1.00} \\
 & GDINO-T        & IR     & \underline{53.11}\std{0.89} & 71.44\std{0.95} & 32.47\std{4.15} & 35.67\std{1.98}  & 40.00\std{1.06} \\
 & \bf MS-GDINO-T & RGB+IR & \textbf{55.69}\std{1.18} & \underline{86.21}\std{1.39} & \textbf{71.84}\std{1.75} & \textbf{70.64}\std{1.61} & \textbf{66.33}\std{0.94} \\
\midrule
\multirow{8}{*}{10-shot}
 & DAMSDet        & RGB+IR & 18.15\std{3.81} & 46.07\std{5.02} & 47.81\std{4.05} & 45.82\std{6.02}  & 38.17\std{1.82} \\
 & CAFF-DINO      & RGB+IR & 10.84\std{4.04} & 31.62\std{6.72} & 51.51\std{6.30} & 29.66\std{8.61}  & 28.92\std{2.63} \\
\cdashline{2-8}
 & YOLOW-M        & RGB    & 44.50\std{0.95} & 83.93\std{0.39} & 68.87\std{3.23} & 63.31\std{4.24}  & 62.27\std{0.84} \\
 & YOLOW-M        & IR     & 50.33\std{1.06} & 69.00\std{0.91} & 54.12\std{4.11} & 29.17\std{5.56}  & 41.75\std{1.13} \\
 & \bf MS-YOLOW-M & RGB+IR & 52.36\std{0.59} & 78.52\std{0.65} & \underline{72.99}\std{6.81} & 59.89\std{3.96} & 62.69\std{0.76} \\
\cdashline{2-8}
 & GDINO-T        & RGB    & 48.19\std{0.64} & \textbf{86.71}\std{0.63} & 66.73\std{1.27} & \underline{70.22}\std{1.10} & \underline{65.61}\std{0.81} \\
 & GDINO-T        & IR     & \underline{54.27}\std{0.39} & 72.61\std{0.98} & 36.73\std{3.68} & 35.35\std{2.53}  & 41.91\std{0.67} \\
 & \bf MS-GDINO-T & RGB+IR & \textbf{55.72}\std{0.72} & \underline{85.82}\std{0.68} & \textbf{73.24}\std{2.29} & \textbf{71.42}\std{2.37} & \textbf{67.16}\std{0.67} \\
\bottomrule
\end{tabular}
}
\caption{Few-shot detection on M3FD. mAP50 reported. Standard deviations over 10 runs shown as subscripts. Bold and underline indicate best and second-best per shot setting. The table presents a truncated subset of categories for readability.}
\label{tab:m3fd_fewshot_perclass}
\end{table*}

\renewcommand{\arraystretch}{1.0}

\subsubsection{Impact of Fusion Design}
\label{sec:fusion_study}

We conduct an in-depth study comparing the three fusion strategies from \cref{sec:fusion_module}, along with the external cons-disc baseline from~\cite{wu2026bridging}. All experiments are performed on the 10-shot FLIR split, keeping all other settings identical to isolate the effect of fusion. Results are reported in \cref{tab:flir_fusion_study}. Across both MS-YOLOW and MS-GDINO, element-wise summation consistently achieves the strongest and most stable performance, reaching 71.15 and 70.25 mAP50 respectively. Channel concatenation with a $1\times1$ convolution substantially underperforms, with high variance across seeds, reflecting severe overfitting 
in the low-data regime. The differential cross-attention module performs competitively but still falls below sum, suggesting that even a principled complementarity-aware design cannot overcome the limited supervision signal. The cons-disc baseline similarly underperforms sum despite its more sophisticated decomposition. These results support our central hypothesis: when VLM semantic 
representations are used, textual supervision already aligns RGB and IR features sufficiently, making complex fusion design unnecessary, particularly under scarce annotation budgets.

\renewcommand{\arraystretch}{1.2}
\begin{table*}[tb]
\centering
\small
\scalebox{1.0}{
\begin{tabular}{@{}llcccc@{}}
\toprule
Model & Fusion & Bicycle & Car & Person & All \\
\midrule
\multirow{4}{*}{MS-YOLOW-M}
 & Sum        & \textbf{59.37}\std{2.92} & \textbf{79.58}\std{1.68} & \textbf{74.60}\std{0.81} & \textbf{71.15}\std{0.59} \\
 & Conv       & 43.97\std{9.20} & 64.48\std{6.38} & 59.93\std{3.95} & 56.14\std{3.07} \\
 & Cross-att  & \underline{54.91}\std{2.74} & \underline{75.64}\std{2.64} & \underline{71.70}\std{1.52} & \underline{67.37}\std{1.48} \\
 & Cons-disc  & 41.06\std{4.56} & 68.11\std{2.17} & 56.86\std{2.82} & 55.25\std{1.13} \\
\cdashline{1-6}
\multirow{4}{*}{MS-GDINO-T}
 & Sum        & \textbf{58.66}\std{3.25} & \textbf{79.88}\std{1.14} & \textbf{74.97}\std{1.26} & \textbf{70.25}\std{1.36} \\
 & Conv       & 48.07\std{6.51} & 72.00\std{1.38} & 63.72\std{3.13} & 61.98\std{2.11} \\
 & Cross-att  & \underline{58.14}\std{2.48} & \underline{77.86}\std{0.82} & \underline{72.10}\std{1.11} & \underline{69.63}\std{1.03} \\
 & Cons-disc  & 50.69\std{3.23} & 73.44\std{1.29} & 64.73\std{1.86} & 62.24\std{2.05} \\
\bottomrule
\end{tabular}
}
\caption{Impact of fusion strategy on FLIR (10-shot). mAP50 reported. 
Standard deviations over 10 runs shown as subscripts. Bold and underline 
indicate best and second-best per model.}
\label{tab:flir_fusion_study}
\end{table*}
\renewcommand{\arraystretch}{1.0}

\subsubsection{Ablation Study on Text Modality}
\label{sec:ablation_text}

We study the role of text supervision in MS-YOLOW and MS-GDINO under 
multiple fusion strategies on FLIR 5-shot (\cref{tab:text_msyolow_ablation}). 
With text, sum fusion achieves the best performance in both models 
(70.69 for MS-YOLOW, 69.88 for MS-GDINO), outperforming more complex 
fusion strategies. Without text, all methods degrade substantially, 
with variance increasing markedly across runs. In MS-YOLOW, the ordering 
partially reverses without text cross-attention (65.73) surpasses 
sum (62.42)  suggesting that without semantic grounding, more 
expressive fusion can partially compensate by exploiting visual 
cross-modal correlations. In MS-GDINO, sum remains best even without 
text (63.25 vs.\ 61.97), indicating this reversal is 
architecture-dependent. Across both models, two conclusions hold 
consistently: (i) text supervision is the dominant performance driver, 
substantially reducing sensitivity to fusion design; and (ii) simple 
sum fusion is the most stable and effective strategy when text 
supervision is available, while complex fusion provides only marginal 
and inconsistent benefits under scarce supervision.

\begin{table}[tb]
\centering
\small
\scalebox{1.0}{
\begin{tabular}{@{}llcc@{}}
\toprule
Model & Fusion & \cmark Text & \xmark Text \\
\midrule
\multirow{3}{*}{MS-YOLOW}
 & Sum       & \textbf{70.69}\std{0.36} & 62.42\std{3.66} \\
 & Cross-att & \underline{66.91}\std{1.16} & \textbf{65.73}\std{2.14} \\
 & Conv      & 58.31\std{7.06} & 49.70\std{5.77} \\
\midrule
\multirow{3}{*}{MS-GDINO}
 & Sum       & \textbf{69.88}\std{0.65} & \textbf{63.25}\std{2.26} \\
 & Cross-att & \underline{68.47}\std{1.44} & \underline{61.97}\std{4.20} \\
 & Concat    & 59.90\std{3.08} & 55.52\std{5.98} \\
\bottomrule
\end{tabular}
}
\caption{Ablation of text supervision and fusion strategy in MS-YOLOW 
and MS-GDINO (5-shot, FLIR). mAP50 reported over 10 runs. Bold and 
underline indicate best and second-best per block.}
\label{tab:text_msyolow_ablation}
\end{table}

\subsubsection{Upper-Bound Training and Scalability}
\label{sec:full_supervision}

We evaluate our frameworks under full supervision to assess upper-bound performance and scalability across model capacities. Results on FLIR and M3FD are summarized in \cref{tab:comparison}. Both frameworks scale consistently with model capacity. On FLIR, MS-GDINO (Swin-L) achieves a new state-of-the-art mAP of 50.5, surpassing DAMSDet (49.3) and Fusion-Mamba (47.0). On M3FD, MS-GDINO (Swin-L) reaches 55.3 mAP, outperforming all prior methods by a significant margin. Notably, these results are obtained with simple sum fusion, further confirming that fusion complexity is not the bottleneck in multispectral object detection.

\begin{table*}[tb]
\centering
\small
\begin{minipage}{0.49\textwidth}
\centering
\setlength{\tabcolsep}{2.5pt}
\scalebox{1}{
\begin{tabular}{lccc}
\toprule
Method & mAP & mAP50 & mAP75 \\
\midrule
CFT~\cite{fang2021cross} & 40.2 & 78.7 & 35.5 \\
ICAFusion~\cite{SHEN2023109913} & 41.4 & 79.2 & 36.9 \\
YOLOXCPCF~\cite{hu2024rethinking} & 44.6 & 82.1 & 41.2 \\
CAFF-DINO*~\cite{helvig2024caff} & 45.6 & 83.1 & 42.3 \\
GMDETR~\cite{xiao2024gm} & 45.8 & 83.9 & 42.6 \\
Fusion-Mamba~\cite{xie2024fusionmamba} & 47.0 & 84.9 & 45.9 \\
DAMSDet~\cite{guo2025damsdet} & 49.3 & \underline{86.6} & 48.1 \\
\midrule
MS-YOLOW-M & 47.8 & 83.5 & 45.9 \\
MS-YOLOW-L & 48.9 & 85.3 & 47.7 \\
MS-GDINO-T & 48.5 & 86.0 & 47.6 \\
MS-GDINO-B & \underline{49.4} & 85.9 & \underline{49.2} \\
MS-GDINO-L & \textbf{50.5} & \textbf{87.8} & \textbf{51.3} \\
\bottomrule
\end{tabular}
}
\end{minipage}
\hfill
\begin{minipage}{0.49\textwidth}
\centering
\setlength{\tabcolsep}{2.5pt}
\scalebox{1}{
\begin{tabular}{lccc}
\toprule
Method & mAP & mAP50 & mAP75 \\
\midrule
CFT~\cite{fang2021cross} & 42.5 & 68.2 & 44.6 \\
ICAFusion~\cite{SHEN2023109913} & 41.9 & 67.8 & 44.5 \\
CAFF-DINO~\cite{helvig2024caff} & 49.0 & 75.5 & 52.7 \\
DAMSDet~\cite{guo2025damsdet} & \underline{52.9} & 80.2 & 56.0 \\
\midrule
MS-YOLOW-M & 48.4 & 72.8 & 51.3 \\
MS-YOLOW-L & 48.7 & 73.0 & 51.1 \\
MS-GDINO-T & 50.6 & 79.0 & 53.5 \\
MS-GDINO-B & 52.5 & \underline{80.7} & \underline{56.1} \\
MS-GDINO-L & \textbf{55.3} & \textbf{83.3} & \textbf{58.4} \\
\bottomrule
\end{tabular}
}
\end{minipage}
\caption{Full supervision comparison on FLIR (left) and M3FD (right). 
Model variants: M (Medium), L (Large), T (Tiny), B (Base). 
Bold and underline indicate best and second-best.}
\label{tab:comparison}
\end{table*}


\subsubsection{From Zero-Shot to Few-Shot}
\label{sec:zero_to_few}
We analyze the progression from zero-shot to few-shot on M3FD 
(\cref{tab:progression}). In the zero-shot setting, both MS-YOLOW and MS-GDINO 
already achieve competitive performance without any task-specific fine-tuning. 
However, a notable gap between RGB-only (63.4, 61.9) and IR-only (35.5, 34.5) 
zero-shot performance suggests that cross-spectral transfer is not yet complete 
at zero shot, and that VLM representations remain biased toward the RGB domain 
they were pretrained on. This gap is dataset-dependent and may reflect the 
particular spectral characteristics of M3FD. Introducing only 5 labeled 
examples yields substantial improvements: MS-GDINO-T improves from 60.6 to 
66.33 mAP50, demonstrating that minimal supervision effectively bridges this 
remaining spectral gap. This progression confirms that (i) VLM semantic 
representations provide a strong initialization for multispectral data, and 
(ii) few-shot fine-tuning is necessary to consolidate cross-spectral alignment, 
particularly on datasets where the RGB-IR domain gap is more pronounced.

\renewcommand{\arraystretch}{1.2}
\begin{table}[tb]
\centering
\small
\scalebox{1.0}{
\begin{tabular}{@{}llccc@{}}
\toprule
Setting & Model & Spectrum & mAP50 \\
\midrule
\multirow{6}{*}{Zero-shot}
 & YOLOW-M    & RGB    & 63.4 \\
 & YOLOW-M    & IR     & 35.5 \\
 & MS-YOLOW-M & RGB+IR & 59.3 \\
 & GDINO-T    & RGB    & 61.9 \\
 & GDINO-T    & IR     & 34.5 \\
 & MS-GDINO-T & RGB+IR & 60.6 \\
\cdashline{1-4}
\multirow{2}{*}{5-shot}
 & MS-YOLOW-M & RGB+IR & 62.19 \\
 & MS-GDINO-T & RGB+IR & 66.33 \\
\bottomrule
\end{tabular}
}
\caption{Zero-shot and 5-shot performance on M3FD. mAP50 reported
over 10 runs.}
\label{tab:progression}
\end{table}
\renewcommand{\arraystretch}{1.0}

\subsubsection{Robustness to Misalignment}
\label{sec:misalign}

We assess robustness to RGB-IR misalignment on FLIR using MS-YOLOW by 
artificially translating the IR modality by $\delta \in \{5, 10, 50, 100\}$ pixels, applied consistently to both training and test splits. As shown in \cref{tab:misalignment_fusion}, all three fusion strategies degrade with increasing shift, but at markedly different rates. Sum fusion is the most robust, retaining 52.88 mAP50 at 100px shift, while conv fusion drops most sharply and cross-attention falls in between. The parameter-free nature of sum fusion prevents overfitting to precise spatial correspondence, making it the most suitable choice for real-world deployment where perfect RGB-IR registration cannot be guaranteed.
\begin{table*}
\centering
\scalebox{1.0}{
\begin{tabular}{@{}lccccc@{}}
\includegraphics[width=0.15\linewidth]{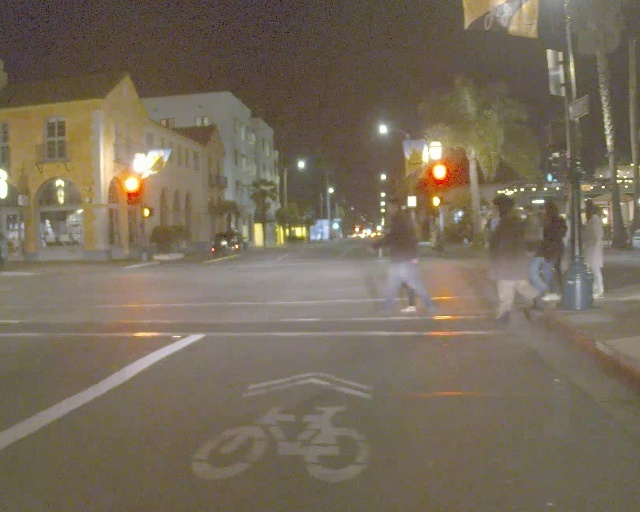} &
\includegraphics[width=0.15\linewidth]{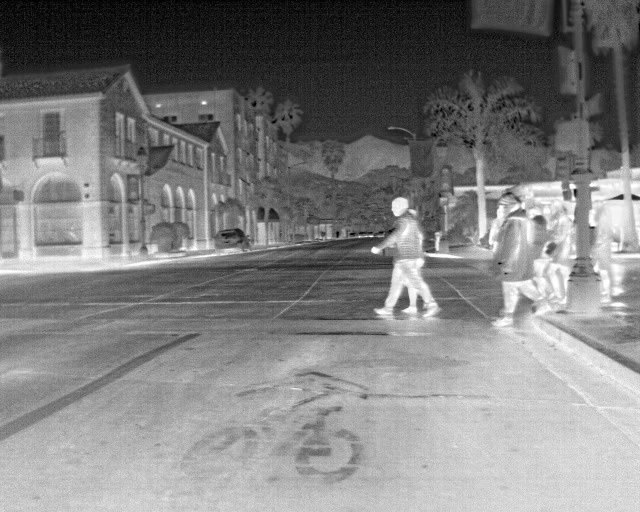} &
\includegraphics[width=0.15\linewidth]{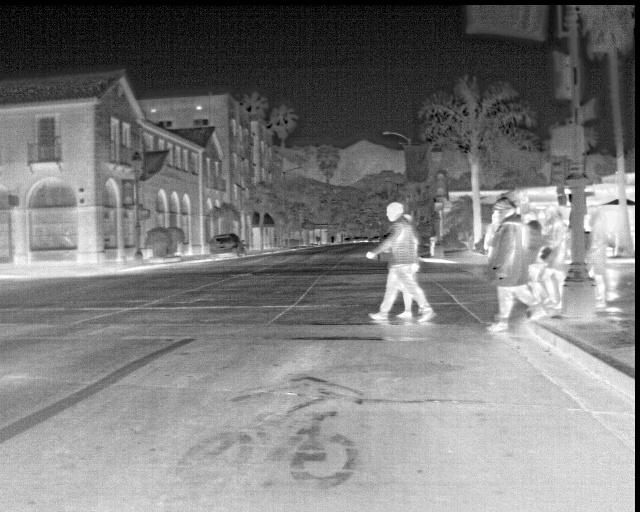} &
\includegraphics[width=0.15\linewidth]{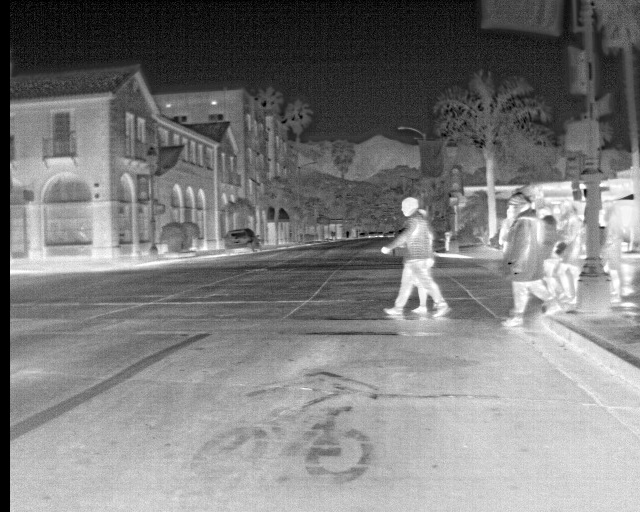} &
\includegraphics[width=0.15\linewidth]{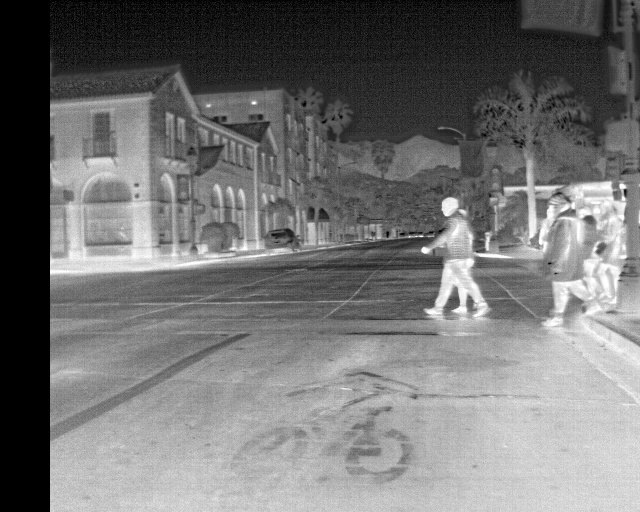} &
\includegraphics[width=0.15\linewidth]{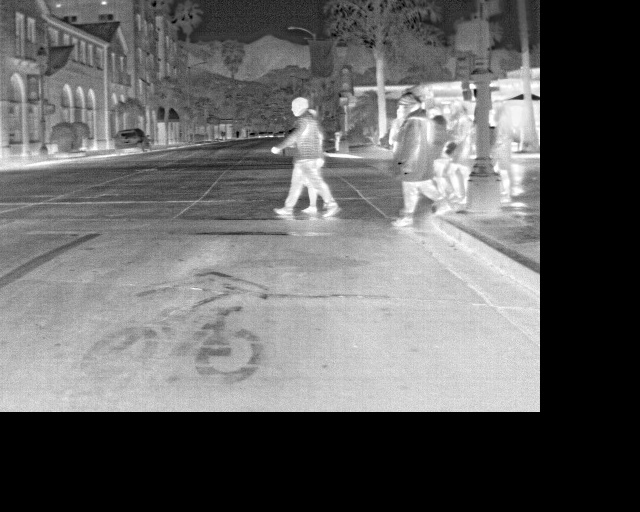} \\
\hline
Fusion & 0px & 5px & 10px  & 50px & 100px  \\
\midrule
\textbf{Sum}   & 71.15 & 64.11 &  56.94 & 53.43 & 52.88 \\
\textbf{Conv} & 56.14 & 46.25 & 41.00 & 40.35 & 41.09\\
\textbf{Cross-att} & 67.00 & 60.06 & 53.36 & 49.47 & 49.90  \\
\bottomrule
\end{tabular}
}
\caption{Robustness of MS-YOLOW to RGB-IR misalignment on M3FD (mAP50 \%). Misalignment is applied to both few-shot train and test sets.}
\label{tab:misalignment_fusion}
\end{table*}

\subsubsection{Effect of Pseudo-Labeling}
\label{sec:tau}

\paragraph{Impact on detection performance.}
Pseudo-labeling was originally applied only to VLM-based methods since their open-vocabulary capacity allows us to  generate accurate labels without task-specific training. For non-VLM baselines, rather than letting them self-generate pseudo-labels, we provide the same VLM-generated pseudo-labels, isolating architectural differences from label source differences.
\cref{tab:flir_pseudo_5shot} shows that VLM-generated pseudo-labels consistently improve accuracy and stability across all methods. On FLIR 5-shot, DAMSDet improves from 40.64 to 54.73 mAP50 and CAFF-DINO from 27.83 to 48.91 mAP50. Crucially, pseudo-labels dramatically reduce seed variance from 3.74 to 0.04 for DAMSDet and from 2.04 to 0.05 for CAFF-DINO. This substantial reduction indicates that semantic pseudo-labels act as a strong regularizer by densifying object-level supervision and reducing optimization ambiguity arising from limited annotations. In practice, sparse few-shot training leads to unstable gradients and strong sensitivity to support set sampling, whereas VLM-generated pseudo-labels provide more consistent supervisory signals across training runs.
\renewcommand{\arraystretch}{1.2}
\begin{table}[tb]
\small
\centering
\scalebox{0.82}{
\begin{tabular}{@{}lcccccc@{}}
\toprule
Method & PL & Bicycle & Car & Person & All \\
\midrule
\multirow{2}{*}{MS-GDINO-T}
 & \xmark & 55.45\std{1.38} & 76.90\std{1.42} & 72.86\std{0.92} & 68.25\std{1.13} \\
 & \cmark & \textbf{57.63}\std{0.69} & \textbf{78.37}\std{0.81} & \textbf{73.70}\std{0.74} & \textbf{69.88}\std{0.65} \\
\midrule
\multirow{2}{*}{DAMSDet}
 & \xmark & 36.48\std{8.23} & 43.12\std{7.49} & 42.32\std{6.26} & 40.64\std{3.74} \\
 & \cmark & \textbf{43.07}\std{0.06} & \textbf{66.23}\std{0.03} & \textbf{54.92}\std{0.06} & \textbf{54.73}\std{0.04} \\
\midrule
\multirow{2}{*}{CAFF-DINO}
 & \xmark & 18.00\std{7.74} & 40.30\std{6.65} & 25.20\std{4.50} & 27.83\std{2.04} \\
 & \cmark & \textbf{29.92}\std{0.07} & \textbf{65.28}\std{0.05} & \textbf{51.50}\std{0.05} & \textbf{48.91}\std{0.05} \\
\bottomrule
\end{tabular}
}
\caption{Effect of VLM-generated pseudo-labels (PL) on FLIR 5-shot. 
mAP50 reported. \cmark~= with pseudo-labels, \xmark~= without.}
\label{tab:flir_pseudo_5shot}
\end{table}
\renewcommand{\arraystretch}{1.0}

\paragraph{Effect of confidence floor $\tau_{\mathrm{floor}}$.}
We further analyze how the confidence floor $\tau_{\text{floor}} \in [0.2, 1.0]$ influences pseudo-label quality and downstream performance. This parameter effectively serves as a bias-variance control: lower values increase recall by allowing more detections to be added as pseudo-labels, while higher values improve precision by filtering out uncertain predictions. As shown in \cref{tab:tau}, moderate floor values around $\tau_{\text{floor}}{=}0.35$ achieve the best overall performance, improving mAP50 by roughly $+2$ points compared to the no-pseudo-label baseline ($\tau_{\text{floor}}{=}1.0$). Lowering the floor to $0.2$ slightly increases recall but introduces noisy pseudo-labels that degrade performance. For high-frequency classes like \emph{Person} and \emph{Car}, such a low threshold maximizes mAP50 by including moderate-confidence true positives, boosting effective training data while adding minimal noise. On the other hand, moderately difficult or ambiguous classes like \emph{Traffic light} and \emph{Motorcycle} require higher thresholds ($\tau_{\text{floor}} = 0.5$) to preserve precision and avoid incorrect pseudo-labels that harm few-shot learning. 

\begin{table}[tb]
\centering
\small
\scalebox{1.0}{
\begin{tabular}{@{}lcccccc c@{}}
\toprule
$\tau_{\mathrm{floor}}$ & Person & Car & Bus & Moto & Tr. Light & Truck & All \\
\midrule
0.20 & \textbf{55.3} & \textbf{86.5} & 69.2 & \underline{68.9} & 46.9 & 60.9 & \underline{64.6} \\
0.35 & \underline{55.2} & \underline{86.1} & \textbf{71.7} & \textbf{70.6} & \underline{51.5} & \textbf{62.7} & \textbf{66.3} \\
0.50 & 51.6 & 82.0 & \underline{70.9} & 65.9 & \textbf{53.0} & \underline{62.0} & 64.3 \\
0.75 & 46.1 & 75.7 & 67.6 & 60.3 & 44.0 & 54.4 & 58.1 \\
1.00 & 46.3 & 71.4 & 69.7 & 58.6 & 50.0 & 50.0 & 57.7 \\
\bottomrule
\end{tabular}
}
\caption{Effect of $\tau_{\mathrm{floor}}$ on MS-GDINO (5-shot, M3FD). 
mAP50 reported. Bold and underline indicate best and second-best per class.}
\label{tab:tau}
\end{table}

\subsubsection{Runtime analysis}
\label{sec:runtime}
\cref{tab:runtime_fps} reveals substantial benefits of multispectral fusion with distinct efficiency-accuracy trade-offs. Mono-spectral RGB baselines show limited performance (22.9-25.3 mAP), demonstrating the fundamental challenge of single-modality detection. The transition to multispectral processing yields dramatic improvements: MS-YOLOW-M achieves +24.9 mAP over its mono-spectral counterpart while maintaining real-time performance (27.9 FPS). YOLOW variants demonstrate superior computational efficiency, with MS-YOLOW-M achieving 27.9 FPS, approximately 3x faster than MS-GDINO-T (8.8 FPS) and 10x faster than MS-GDINO-L (2.6 FPS). However, MS-GDINO-L achieves the best detection performance with 50.5 mAP and 87.8 mAP50, representing a +25.2 mAP improvement over its mono-spectral baseline. This analysis reveals clear deployment guidelines: MS-YOLOW for real-time applications requiring efficiency, and MS-GDINO for accuracy-critical scenarios. Both multispectral approaches substantially outperform mono-spectral counterparts, validating our core premise.

\begin{table}[tb]
\centering
\small
\setlength{\tabcolsep}{2.5pt} 
\scalebox{1.0}{
\begin{tabular}{lccccc}
\toprule
Model & Spec. & Params (M) & FPS & mAP & mAP$_{50}$ \\
\midrule
YOLOW-M & RGB  & 92  & 31.6 & 22.9 & 55.2 \\
YOLOW-L & RGB  & 110 & 28.2 & 23.2 & 55.6 \\
GDINO-T & RGB  & 172  & 11.9 & 22.6 & 54.2 \\
GDINO-L & RGB  & 341  & 3.5 & 25.3 & 60.2 \\
\midrule
MS-YOLOW-M & RGB+IR & 104 & 27.9 & 47.8 & 83.5 \\
MS-YOLOW-L & RGB+IR & 131 & 24.0 & \underline{48.9} & 85.3 \\
MS-GDINO-T & RGB+IR & 198 & 8.8 & 48.5 & \underline{86.0} \\
MS-GDINO-L & RGB+IR & 367 & 2.6 & \textbf{50.5} & \textbf{87.8} \\
\bottomrule
\end{tabular}
}
\caption{Runtime and detection performance on FLIR using an A100 GPU at $640{\times}640$ input. Best and second-best are in \textbf{bold} and \underline{underline}. The \textbf{Sum} fusion was used for the experiments.}
\label{tab:runtime_fps}
\end{table}

\section{Conclusion and Future Work}
\label{sec: conclusion}
In this work, we presented the first systematic study of VLMs for few-shot multispectral object detection, bridging the gap between language-grounded pretraining and multispectral perception. By introducing lightweight adaptations of \textbf{Grounding DINO} and \textbf{YOLO-World}, we demonstrated that semantic priors learned from large-scale image-text data effectively transfer across spectral domains, substantially improving data efficiency and robustness. Extensive experiments on the FLIR and M3FD benchmarks established new state-of-the-art results in both few-shot and fully supervised regimes, while further experiments revealed that adaptive pseudo-labeling and simple fusion mechanisms are sufficient to unlock strong cross-spectral generalization. Our findings highlight that semantic grounding, rather than complex fusion, is the key to robust multispectral learning. The success of VLMs in this setting suggests that modality-agnostic semantic spaces learned from web-scale data can serve as universal representations for heterogeneous sensor modalities. A promising direction of research is to study the \textbf{zero-shot adaptability of VLMs to rare or unseen classes in multispectral domains}, as current datasets primarily contain common categories such as \emph{person} and \emph{car}. Evaluating and improving transfer to uncommon or fine-grained categories would provide deeper insight into the limits of semantic generalization across wavelengths.

{
    \small
    \bibliographystyle{ieeenat_fullname}
    \bibliography{BiblioManu}
}

 \clearpage
\setcounter{page}{1}
\maketitlesupplementary

This supplementary material provides additional details and results 
complementing the main paper. We first detail the adaptive 
pseudo-labeling strategy and its confidence calibration analysis 
(\cref{sec:pseudo_labeling_threshold}). We then provide full 
implementation details of the fusion paradigms studied in the main paper 
(\cref{sec:fusion_para}), including the differential cross-attention 
module and the text-guided consensus-discrepancy baseline. Extended 
experimental results are presented in \cref{app:additional_results}, 
organized as follows: 30-shot scaling experiments on FLIR and M3FD 
(\cref{app:additional_results_30_shots}); cross-dataset evaluation on 
DroneVehicle (\cref{app:dronevehicle}); an analysis of open-set generalization under multispectral fine-tuning (\cref{app:open_set_generalization}); a direct comparison with FSMODNet (\cref{sec:fsmodnet_comp}) and an analysis of the sum vs.\ $1{\times}1$ conv fusion gap (\cref{sec:random}) and 
qualitative detection visualizations under challenging conditions 
(\cref{sec:visu}). Unless otherwise stated, all experiments follow the 
same data splits, training protocols, hyperparameters, and evaluation 
metrics as the main paper.

\section{Pseudo-Labelling threshold}
\label{sec:pseudo_labeling_threshold}

\begin{figure}[tb]
    \centering
    \includegraphics[width=0.5\textwidth]{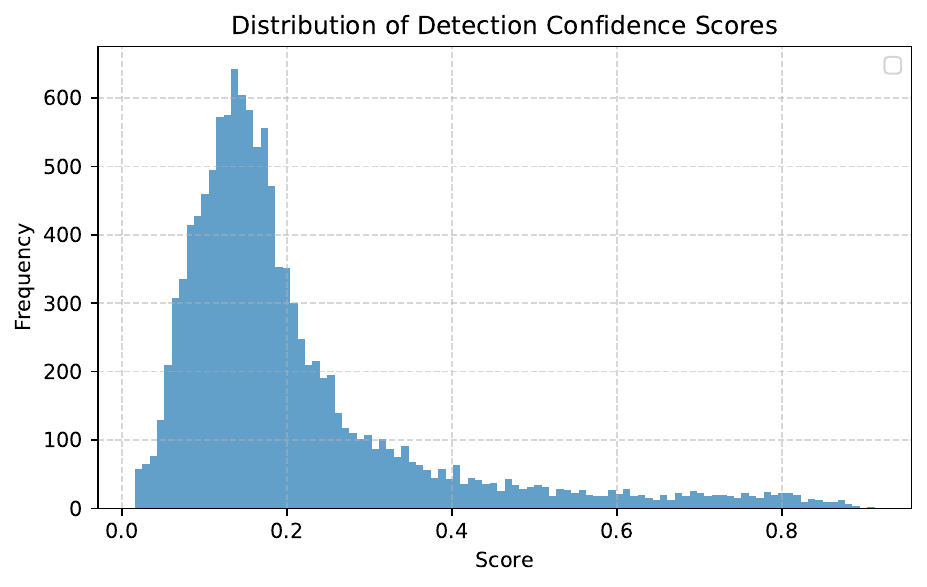}
    \includegraphics[width=0.5\textwidth]{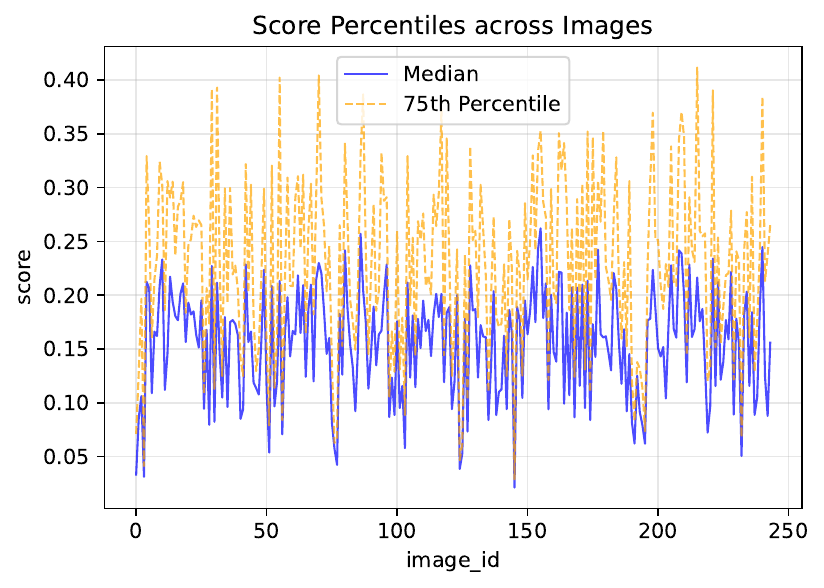}
    \caption{
    Analysis of detection confidence scores from Grounding DINO on few-shot training splits.
    \textbf{Top:} Histogram of all prediction scores shows a right-skewed distribution with high frequency of low-confidence detections.
    \textbf{Bottom:} Per-image statistics showing median and 75th percentile scores reveal substantial variation across images, with some having consistently high-confidence predictions while others contain predominantly low-scoring detections.
    }
    \label{fig:score_distribution}
\end{figure}

To generate reliable pseudo-labels for augmenting the few-shot datasets, we analyze the confidence calibration of a reference model (Grounding DINO). The goal is to automatically identify high-confidence detections that likely correspond to unlabeled true objects, while filtering out uncertain predictions that could introduce noise. We begin by evaluating the model on the 5-shot training splits of M3FD and collecting the confidence scores of the top-50 predictions per image. As shown in \cref{fig:score_distribution}, the overall distribution of prediction scores exhibits a right-skewed pattern, with a high frequency of detections at lower confidence levels ($\text{scores} \leq 0.3 $), indicating that a substantial portion of the model's predictions are low-confidence and potentially unreliable. The frequency of predictions decreases rapidly as confidence increases, suggesting that high-confidence detections are relatively rare. \cref{fig:score_distribution} also summarizes per-image confidence distributions by showing the median and 75th percentile of the top-50 predictions for each image. For example, if the 75th percentile for an image is 0.3, this means that 38 out of the 50 top predictions have confidence scores below or equal to 0.3, while only the remaining 12 predictions exceed this value. The substantial vertical spread across images indicates dramatic inter-image variability: some images have consistently high-confidence detections, whereas for other images even the top quartile of predictions falls below 0.1 confidence. This variability demonstrates that applying a fixed global confidence threshold would be suboptimal, as the reliability of pseudo-labels depends heavily on individual image content and context. Such under-confidence is consistent with the limited supervision in few-shot settings, where many valid objects remain unannotated, causing the model to underestimate its detection certainty. As described in the main text, we address this issue using an adaptive thresholding strategy when generating pseudo-labels, selecting high-confidence predictions in a manner tailored to each image’s score distribution. For completeness, we restate the formulation here. For each image, the threshold $\tau$ is computed as
\[
\tau = \max(\mu + \sigma, \tau_{\text{floor}}),
\]
where $\mu$ and $\sigma$ are the mean and standard deviation of the detection scores within that image, and $\tau_{\text{floor}}$ defines a confidence floor. The adaptive term $\mu + \sigma$ adjusts to local score distributions, promoting higher selectivity for well-calibrated images and tolerance for uncertain ones. The floor $\tau_{\text{floor}}$ serves as a safety guard, preventing the threshold from collapsing to unreasonably low values when the model is highly uncertain. Pseudo-labels are then selected from detections with $s_j \ge \tau$ and a maximum intersection-over-union (IoU) below 0.3 with any ground-truth annotation of the same class. This ensures that pseudo-labels complement, rather than duplicate, existing annotations. The resulting pseudo-labeled datasets are stored as extended COCO-format annotation files and used for subsequent fine-tuning experiments.

\section{Implementation Details of the Compared Fusion Paradigms}
\label{sec:fusion_para}

\subsection{Differential Cross-Attention (cross-att)}
\label{sec:MDCA}
We explore a differential cross-attention formulation for multimodal fusion. Unlike standard cross-attention, which aggregates all cross-modal interactions uniformly, differential attention~\cite{ye2025differential} contrasts multiple attention responses to suppress correlated activations and emphasize complementary information across modalities. The module is bidirectional, with each modality serving as query for one stream and key/value for the other. For simplicity, we present only RGB $\rightarrow$ IR direction. We compute a differential cross-attention from RGB queries to IR keys/values. Specifically, RGB features are projected into queries $(Q_1, Q_2)$, while IR features are projected into keys $(K_1, K_2)$ and values $V_{\mathrm{IR}}$, with all projection matrices learned. The RGB-refined output is:
\begin{align}
\mathbf{O}_{\mathrm{RGB}} =
\left(
\mathrm{softmax}\!\left(\frac{Q_1 K_1^\top}{\sqrt{d}}\right)
-
\lambda\, \mathrm{softmax}\!\left(\frac{Q_2 K_2^\top}{\sqrt{d}}\right)
\right)
V_{\mathrm{IR}}.
\end{align}

where the differential coefficient is defined as:
\begin{align}
\lambda = e^{\langle \lambda_{q_1}, \lambda_{k_1} \rangle}
-
e^{\langle \lambda_{q_2}, \lambda_{k_2} \rangle} \cdot \lambda_{\mathrm{init}},
\end{align}
with $\lambda_{q_i}, \lambda_{k_i} \in \mathbb{R}^{d}$ learnable vectors and $\lambda_{\mathrm{init}} \in (0,1)$ a fixed scalar. The subtraction suppresses shared attention responses, emphasizing cross-modal discrepancies.

\paragraph{Linear Approximation.}
The exact formulation above incurs $\mathcal{O}(N^2)$ complexity, which is prohibitive for dense spatial feature maps. We therefore replace the softmax kernel with $\phi(x)=\mathrm{ELU}(x)+1$~\cite{katharopoulos2020transformers}. The resulting linearized RGB output is:
\begin{align}
\mathbf{O}_{\mathrm{RGB}} =
\frac{\phi(Q_1)\left(\phi(K_1)^\top V_{\mathrm{IR}}\right)}{\phi(Q_1)\mathbf{1}^\top\phi(K_1)}
-
\lambda\,
\frac{\phi(Q_2)\left(\phi(K_2)^\top V_{\mathrm{IR}}\right)}{\phi(Q_2)\mathbf{1}^\top\phi(K_2)}.
\end{align}
The differential cross-attention outputs are refined using RMSNorm~\cite{zhang2019rmsnorm}, residual connections, and a SwiGLUFFN~\cite{shazeer2020glu}:
\begin{align}
\mathcal{G}'_{m} 
    &= (1 - \lambda_{\mathrm{init}})
       \cdot
       \mathrm{RMSNorm}\!\left(\mathbf{O}_m\right)
       + \mathcal{F}'_{m}, \\
\tilde{\mathcal{F}}_{m}
    &= \mathrm{SwiGLU}\!\left(\mathcal{G}'_{m}\right)
       + \mathcal{G}'_{m},
\end{align}
where $m \in \{\mathrm{RGB}, \mathrm{IR}\}$ and $(1-\lambda_{\mathrm{init}})$ stabilizes training by controlling the magnitude of the differential signal following~\cite{ye2025differential}. The final fused representation is obtained as:
\begin{align}
\mathbf{F}_{\mathrm{fused}}
=
\tilde{\mathcal{F}}_{\mathrm{RGB}}
+
\tilde{\mathcal{F}}_{\mathrm{IR}}.
\end{align}

\paragraph{Complexity and Runtime Comparison.}
\cref{tab:cross_att_complexity} compares the standard and linearized variants 
of the differential cross-attention module in terms of GFLOPs, and inference speed.  The standard formulation incurs $\mathcal{O}(N^2 d)$ complexity, scaling 
quadratically with the number of spatial tokens $N$ — prohibitive for dense 
feature maps at detection resolution. The linear approximation reduces this to 
$\mathcal{O}(N d^2)$ by factorizing the attention computation via context 
matrices, yielding a $4.5\times$ reduction in GFLOPs (68.15 vs 15.31) and a 
$1.24\times$ improvement in throughput (13.02 vs 16.10 FPS), with only a 
marginal drop in mAP50 (67.90 vs 67.00, $-0.90$ points) 
\begin{table}[tb]
\centering
\small
\scalebox{0.9}{
\begin{tabular}{@{}lccccc@{}}
\toprule
Variant & Complexity & GFLOPs & FPS & mAP50 \\
\midrule
Cross-att (standard) & $\mathcal{O}(N^2 d)$ & 68.15 & 13.02 & 67.90 \\
Cross-att (linear)   & $\mathcal{O}(N d^2)$ & 15.31 & 16.10 & 67.00 \\
\bottomrule
\end{tabular}
}
\caption{Comparison of computational complexity and inference speed between standard and linearized differential cross-attention on the FLIR dataset (10-shot setting, MS-YOLO-W-M, $640{\times}640$ input, NVIDIA RTX A6000). GFLOPs are computed for the fusion module only, isolating cross-modal interaction cost.}
\label{tab:cross_att_complexity}
\end{table}

\subsection{Text-Guided Consensus-Discrepancy Fusion}
\label{sec:consensus}

We additionally evaluate the text-guided consensus-discrepancy fusion 
of~\cite{wu2026bridging} as an external baseline. This method uses text as a 
shared semantic bridge to align RGB and IR responses under a unified category 
condition. For each modality $m$, text-visual affinity maps are computed by 
matching enhanced visual features against text embeddings:
\begin{align}
A_m = \sigma\!\left(\mathcal{F}'_m \cdot (T'_m)^\top\right) 
\in \mathbb{R}^{N \times N_t},
\end{align}
where $\sigma$ is the sigmoid function. The RGB-IR interaction is decomposed into 
consensus and discrepancy support patterns:
\begin{align}
M_{\mathrm{cons}} &= A_{\mathrm{RGB}} \odot A_{\mathrm{IR}}, \\
M_{\mathrm{disc}} &= A_{\mathrm{RGB}} \odot (1 - A_{\mathrm{IR}}) 
                   + A_{\mathrm{IR}} \odot (1 - A_{\mathrm{RGB}}),
\end{align}
where $\odot$ denotes element-wise multiplication. The IR features are then 
recalibrated using both support patterns via learnable modulation coefficients 
$\alpha$ and $\beta$:
\begin{align}
\tilde{\mathcal{F}}_{\mathrm{IR}} = \mathcal{F}_{\mathrm{IR}} \odot 
(1 + \beta M_{\mathrm{cons}}) \odot (1 + \alpha M_{\mathrm{disc}}),
\end{align}
where $\beta M_{\mathrm{cons}}$ reinforces stable cross-modal consensus responses, 
while $\alpha M_{\mathrm{disc}}$ acts as a learned gate on discrepancy support, 
preserving potentially discriminative cues while suppressing discrepancy-induced 
errors. The recalibrated IR features are then aligned to the RGB channel space 
via a learned projection $\psi$, generating a spatial attention map 
$W_{\mathrm{att}} = \phi(\psi(\tilde{\mathcal{F}}_{\mathrm{IR}}))$ that 
reweights RGB features. The final fused representation is:
\begin{align}
\hat{\mathcal{F}}_{\mathrm{RGB}} &= \mathcal{F}_{\mathrm{RGB}} \odot W_{\mathrm{att}}, \\
\mathbf{F}_{\mathrm{fused}} &= \mathrm{Conv}\!\left(
    \mathrm{Concat}\!\left(\hat{\mathcal{F}}_{\mathrm{RGB}},\, 
    \psi(\tilde{\mathcal{F}}_{\mathrm{IR}})\right)
\right).
\end{align}

\section{Additional Experimental Results}
\label{app:additional_results}

\subsection{30-shot Results}
\label{app:additional_results_30_shots}

To further analyze performance scaling with increased supervision, we extend our few-shot experiments to the 30-shot regime on both the FLIR and M3FD benchmarks. Results are summarized in \cref{tab:flir_30shot_sum,tab:m3fd_30shot_sum}.  

On \textbf{FLIR}, both multispectral frameworks exhibit steady improvements compared to their 10-shot counterparts. MS-YOLOW achieves 72.05~mAP$_{50}$, while MS-GDINO reaches 71.52~mAP$_{50}$, marking absolute gains of +0.9 and +1.3 points, respectively. Class-wise analysis shows that underrepresented categories such as \textit{Bicycle} benefit the most from additional annotations (\(+2.95\)~mAP$_{50}$ for MS-YOLOW), whereas already saturated classes like \textit{car} and \textit{person} experience marginal but consistent improvements.  

A similar trend appears on \textbf{M3FD}, where both frameworks maintain strong cross-domain performance. MS-YOLOW improves from 62.69 to 64.44~mAP, while MS-GDINO increases from 67.01 to 67.57~mAP, demonstrating robust scaling despite distributional and spectral shifts. Notably, MS-GDINO continues to excel on complex classes such as \textit{bus} and \textit{motorcycle}, reflecting the advantage of transformer-based multimodal reasoning, whereas MS-YOLOW sustains higher throughput and competitive per-class detection, particularly on \textit{car} and \textit{person}.  

Overall, the 30-shot results across both datasets confirm that the proposed multispectral frameworks remain data-efficient and generalizable, with consistent, class-dependent improvements. The incremental gains highlight the complementary strengths of each architecture, YOLOW favoring efficiency and dense supervision, while GDINO excels in semantic grounding and multimodal fusion under low-data conditions.

\renewcommand{\arraystretch}{1.2}

\begin{table*}[tb]
\centering
\small
\setlength{\tabcolsep}{3.5pt} 

\scalebox{1}{
\begin{tabular}{@{}llcccccc@{}}
\toprule
Shots & Method & Spectrum & Bicycle & Car & Person & All \\
\midrule
\multirow{8}{*}{10-shot} 
 & DAMSDet   & RGB+IR & 39.69\std{11.40} & 50.31\std{8.20} & 52.19\std{4.40} & 47.40\std{5.09} \\
 & CAFF-DINO & RGB+IR & 24.30\std{7.36}  & 40.50\std{4.69} & 31.90\std{4.71} & 32.23\std{1.78} \\
 \cdashline{2-7}

 & YOLOW-M   & RGB    & 43.14\std{2.16} & 69.18\std{1.16} & 56.25\std{1.89} & 56.18\std{1.08} \\
 & YOLOW-M   & IR     & 56.85\std{1.73} & 77.92\std{0.64} & \textbf{76.65}\std{0.99} & 69.81\std{0.77} \\
 & MS-YOLOW-M & RGB+IR & \textbf{59.37}\std{2.92} & \underline{79.58}\std{1.68} & \underline{74.60}\std{0.81} & \textbf{71.15}\std{0.59} \\
 \cdashline{2-7}

 & GDINO-T   & RGB    & 44.84\std{1.18} & 70.59\std{1.20} & 59.37\std{1.08} & 58.27\std{0.88} \\
 & GDINO-T   & IR     & 51.98\std{1.86} & 76.79\std{0.55} & 74.69\std{0.85} & 67.82\std{0.80} \\
 & MS-GDINO-T & RGB+IR & \underline{58.66}\std{3.25} & \textbf{79.88}\std{1.14} & 74.97\std{1.26} & \underline{70.25}\std{1.36} \\
\midrule
\multirow{8}{*}{30-shot} 
 & DAMSDet & RGB+IR & 49.88\std{8.40} & 58.39\std{8.45} & 62.98\std{4.69} & 57.08\std{5.09} \\
 & CAFF-DINO & RGB+IR & 35.82\std{3.22} & 51.14\std{3.72} & 46.42\std{4.72} & 44.46\std{2.74} \\
 \cdashline{2-7}
 & YOLOW-M & RGB & 46.06\std{2.44} & 69.72\std{2.77} & 56.90\std{1.63} & 57.56\std{1.58} \\
 & YOLOW-M & IR & 55.84\std{2.66} & 77.70\std{1.50} & \textbf{76.39}\std{1.26} & 69.98\std{0.75} \\
 & \textbf{MS-YOLOW-M }& RGB+IR & \textbf{62.32}\std{2.62} & \underline{79.16}\std{1.18} & 74.67\std{0.83} & \textbf{72.05}\std{0.62} \\
 \cdashline{2-7}
 & GDINO-T & RGB & 47.09\std{2.33} & 72.34\std{0.71} & 60.97\std{0.92} & 60.13\std{0.94} \\
 & GDINO-T & IR & 52.27\std{2.01} & 79.18\std{1.10} & 76.05\std{0.93} & 69.50\std{0.82} \\
 & \textbf{MS-GDINO-T} & RGB+IR & \underline{58.30}\std{1.89} & \textbf{79.62}\std{0.93} & \underline{74.68}\std{1.18} & \underline{70.87}\std{1.09} \\
\bottomrule
\end{tabular}
}
\caption{10-shot and 30-shot object detection results on the FLIR dataset. MS-YOLOW and MS-GDINO are reported using sum fusion. Best and second-best results are highlighted in \textbf{bold} and \underline{underline}, respectively.}
\label{tab:flir_30shot_sum}
\end{table*}

\begin{table*}[tb]
\centering
\small
\setlength{\tabcolsep}{3.5pt} 
\scalebox{1}{
\begin{tabular}{@{}llccccccc@{}}
\toprule
Shots & Method & Spectrum & Person & Car & Bus & Motorcycle & All \\
\midrule
\multirow{8}{*}{10-shot}
 & DAMSDet        & RGB+IR & 18.15\std{3.81} & 46.07\std{5.02} & 47.81\std{4.05} & 45.82\std{6.02}  & 38.17\std{1.82} \\
 & CAFF-DINO      & RGB+IR & 10.84\std{4.04} & 31.62\std{6.72} & 51.51\std{6.30} & 29.66\std{8.61}  & 28.92\std{2.63} \\
\cdashline{2-8}
 & YOLOW-M        & RGB    & 44.50\std{0.95} & 83.93\std{0.39} & 68.87\std{3.23} & 63.31\std{4.24}  & 62.27\std{0.84} \\
 & YOLOW-M        & IR     & 50.33\std{1.06} & 69.00\std{0.91} & 54.12\std{4.11} & 29.17\std{5.56}  & 41.75\std{1.13} \\
 & \bf MS-YOLOW-M & RGB+IR & 52.36\std{0.59} & 78.52\std{0.65} & \underline{72.99}\std{6.81} & 59.89\std{3.96} & 62.69\std{0.76} \\
\cdashline{2-8}
 & GDINO-T        & RGB    & 48.19\std{0.64} & \textbf{86.71}\std{0.63} & 66.73\std{1.27} & \underline{70.22}\std{1.10} & \underline{65.61}\std{0.81} \\
 & GDINO-T        & IR     & \underline{54.27}\std{0.39} & 72.61\std{0.98} & 36.73\std{3.68} & 35.35\std{2.53}  & 41.91\std{0.67} \\
 & \bf MS-GDINO-T & RGB+IR & \textbf{55.72}\std{0.72} & \underline{85.82}\std{0.68} & \textbf{73.24}\std{2.29} & \textbf{71.42}\std{2.37} & \textbf{67.16}\std{0.67} \\
\midrule
\multirow{8}{*}{30-shot}
 & DAMSDet & RGB+IR & 23.25\std{4.04} & 56.43\std{4.20} & 56.39\std{4.75} & 53.39\std{3.88} & 46.56\std{1.12} \\
 & CAFF-DINO & RGB+IR & 13.85\std{3.47} & 40.68\std{4.60} & 55.82\std{6.61} & 42.27\std{6.13} & 36.35\std{2.04} \\
 \cdashline{2-8}
 & YOLOW-M & RGB & 43.63\std{0.40} & 76.90\std{3.80} & 70.70\std{2.93} & 65.18\std{2.03}  & 62.12\std{1.39} \\
 & YOLOW-M & IR & 45.96\std{2.56} & 56.96\std{5.36} & 58.82\std{2.91} & 34.02\std{2.79} & 40.61\std{0.51} \\
 & \textbf{ MS-YOLOW-M}& RGB+IR & 52.04\std{1.30} & 85.96\std{2.39} & \underline{73.24}\std{4.61} & 62.61\std{3.13} & 64.44\std{0.69} \\
 \cdashline{2-8}
 & GDINO-T & RGB & 48.71\std{0.30} & \textbf{87.30}\std{0.24} & 68.19\std{1.95} & \underline{69.81}\std{1.11} & \underline{65.94}\std{0.71} \\
 & GDINO-T & IR & \underline{55.51}\std{1.00} & 74.92\std{0.45} & 47.89\std{3.27} & 39.63\std{1.58} & 46.92\std{1.11} \\
 & \textbf{ MS-GDINO-T} & RGB+IR & \textbf{56.24}\std{0.52} & \underline{85.97}\std{0.68} & \textbf{74.10}\std{1.59} & \textbf{72.10}\std{0.96} & \textbf{67.55}\std{0.95} \\
\bottomrule
\end{tabular}
}
\caption{10-shot and 30-shot object detection results on the M3FD dataset. MS-YOLOW and MS-GDINO are reported using sum fusion. Best and second-best results are highlighted in \textbf{bold} and \underline{underline}, respectively.}
\label{tab:m3fd_30shot_sum}
\end{table*}
\renewcommand{\arraystretch}{1.0}

\subsection{Experiments on DroneVehicle}
\label{app:dronevehicle}

To further demonstrate the generalizability of our approach, we evaluate the proposed MS-GDINO-T on the \textbf{DroneVehicle}~\cite{sun2020drone} dataset under a 10-shot few-shot setting. The dataset contains aerial RGB-IR images with several vehicle categories. Standard preprocessing is applied to adapt oriented bounding boxes to axis-aligned detection. Table~\ref{tab:dronevehicle_fewshot_fusion} reports per-class mAP$_{50}$ results. MS-GDINO-T consistently outperforms both single-modality baselines and the specialized multispectral detector DAMSDet. Compared to RGB-only GDINO-T, our multispectral formulation improves overall performance from 14.60 to 18.46 mAP$_{50}$, corresponding to a relative gain of more than 26\%. Similar improvements are observed across all categories, with particularly notable gains for the dominant \emph{Car} class (+13.85 points) and the \emph{Freight Car} category (+1.65 points). These results demonstrate that the benefits of vision-language transfer extend beyond ground-based multispectral benchmarks and generalize to aerial vehicle detection scenarios. The consistent improvements across datasets reinforce the conclusion that semantic alignment provided by pretrained VLMs enables effective and data-efficient multispectral perception.

\renewcommand{\arraystretch}{1.2}

\begin{table*}[tb]
\small
\centering
\setlength{\tabcolsep}{2.5pt} 
\scalebox{1.0}{
\begin{tabular}{@{}llcccccccc@{}}
\toprule
Shots & Method & Spectrum & Car & Truck & Bus & Van & Fr. Car & All \\
\midrule
\multirow{8}{*}{5} 
 & DAMSDet        & RGB+IR & 17.69\std{4.52} & 1.70\std{0.52} & 5.72\std{1.89} & 0.83\std{0.36} & 1.60\std{0.84} & 5.48\std{0.98} \\
 & CAFF-DINO      & RGB+IR & 11.25\std{5.72} & 1.89\std{0.36} & 9.44\std{4.20} & 1.44\std{0.64} & 2.43\std{0.94} & 5.39\std{1.25} \\
\cdashline{2-9}
 & YOLOW-M        & RGB    & 33.63\std{2.37} & 1.85\std{0.58} & 2.45\std{1.47} & 0.39\std{0.12} & 1.44\std{0.66} & 8.04\std{0.60} \\
 & YOLOW-M        & IR     & 25.05\std{6.31} & 1.64\std{0.69} & 4.55\std{3.07} & 0.37\std{0.18} & 1.29\std{0.54} & 6.59\std{1.52} \\
 & \bf MS-YOLOW-M & RGB+IR & 46.70\std{2.91} & 3.30\std{1.43} & 7.60\std{4.47} & 1.43\std{0.84} & 1.50\std{0.80} & 12.20\std{1.38} \\
\cdashline{2-9}
 & GDINO-T        & RGB    & 47.70\std{5.62} & \textbf{7.54}\std{2.05} & \textbf{13.38}\std{4.22} & \underline{2.29}\std{0.62} & \underline{3.51}\std{0.98} & 14.95\std{2.47} \\
 & GDINO-T        & IR     & \underline{57.08}\std{6.57} & 4.72\std{1.25} & \underline{12.65}\std{3.90} & 2.16\std{0.45} & 3.09\std{0.91} & \textbf{16.78}\std{2.07} \\
 & \bf MS-GDINO-T & RGB+IR & \textbf{57.10}\std{3.76} & \underline{6.91}\std{1.19} & 10.40\std{1.50} & \textbf{2.70}\std{0.31} & \textbf{4.10}\std{0.62} & \underline{16.28}\std{1.04} \\
\midrule
\multirow{8}{*}{10} 
 & DAMSDet        & RGB+IR & 22.54\std{4.76} & 2.19\std{0.77} & 7.68\std{4.15} & 1.57\std{0.56} & 3.06\std{0.53} & 7.41\std{0.72} \\
 & CAFF-DINO      & RGB+IR & 13.41\std{6.24} & 2.45\std{0.40} & 9.25\std{3.25} & 1.73\std{0.65} & 3.57\std{0.66} & 6.09\std{1.53} \\
\cdashline{2-9}
 & YOLOW-M        & RGB    & 36.32\std{2.07} & 4.38\std{0.75} & 5.20\std{1.39} & 1.37\std{0.12} & 1.63\std{0.42} & 9.78\std{0.32} \\
 & YOLOW-M        & IR     & 32.70\std{5.13} & 2.36\std{0.51} & 7.98\std{3.02} & 1.59\std{0.14} & 1.65\std{0.50} & 9.24\std{1.34} \\
 & \bf MS-YOLOW-M & RGB+IR & 49.31\std{4.41} & 4.36\std{1.83} & 9.27\std{4.57} & 1.83\std{0.79} & 2.37\std{0.67} & 13.43\std{1.88} \\
\cdashline{2-9}
 & GDINO-T        & RGB    & \underline{54.33}\std{2.69} & \underline{5.89}\std{1.42} & \underline{13.55}\std{4.33} & \underline{2.46}\std{0.} & \underline{3.96}\std{0.98} & \underline{15.65}\std{1.18} \\
 & GDINO-T        & IR     & 47.93\std{6.39} & 5.10\std{1.51} & 11.97\std{3.51} & 2.30\std{0.59} & 3.84\std{1.16} & 14.90\std{2.62} \\
 & \bf MS-GDINO-T & RGB+IR & \textbf{60.85}\std{2.38} & \textbf{8.15}\std{2.03} & \textbf{14.33}\std{4.62} & \textbf{2.89}\std{0.43} & \textbf{5.05}\std{0.64} & \textbf{18.26}\std{1.63} \\
\bottomrule
\end{tabular}
}
\caption{Few-shot detection results on DroneVehicle. ``All'' reports mAP50 across 
all classes. Standard deviations over 10 runs shown as subscripts where available. 
Bold and underline indicate best and second-best.}
\label{tab:dronevehicle_fewshot_fusion}
\end{table*}
\renewcommand{\arraystretch}{1.0}

\subsection{Open-Set Generalization in Multispectral Fine-Tuning}
\label{app:open_set_generalization}

VLMs naturally support open-set recognition, allowing them to localize categories unseen during training. A common concern, however, is that domain-specific fine-tuning may reduce this generalization ability by overfitting to the target domain. To examine whether multispectral fine-tuning preserves cross-domain transfer, we fine-tune Grounding DINO on FLIR and evaluate on the unseen M3FD benchmark. As shown in \cref{tab:open_set_m3fd}, the RGB-only zero-shot model obtains 38.9~mAP and 61.9~mAP$_{50}$ on M3FD, substantially outperforming the IR-only zero-shot model (19.8~mAP, 34.5~mAP$_{50}$), which suffers from a larger domain gap. 
With only 5 labeled FLIR images, the multispectral model achieves 38.9~mAP and 63.3~mAP$_{50}$, matching or slightly surpassing the RGB zero-shot performance. Increasing the supervision to 30 images yields 37.6~mAP and 60.6~mAP$_{50}$, still close to the zero-shot baseline. In contrast, full-shot fine-tuning reduces performance to 26.5~mAP and 48.4~mAP$_{50}$, indicating that heavy specialization to FLIR reduces open-set transfer. Overall, these results show that multispectral fine-tuning preserves open-set generalization under low-shot supervision and that the degradation observed under full-shot fine-tuning is primarily due to dataset-specific overfitting rather than catastrophic forgetting. These observations motivate a deeper examination of the zero-shot adaptability of VLMs in multispectral settings, particularly their ability to generalize to object categories that are rare or entirely unseen during fine-tuning.

\begin{table}[tb]
\centering
\small
\scalebox{1}{
\begin{tabular}{lcccc}
\toprule
  Setting & Spectrum & mAP & mAP50 \\
\midrule
Zero-shot transfer & RGB & 38.9 & 61.9 \\
Zero-shot transfer & IR & 19.8 & 34.5 \\
Fine-tuned (FLIR -- 5 shot) & RGB + IR & 38.9  & 63.3  \\
Fine-tuned (FLIR -- 30 shot) & RGB + IR & 37.6  & 60.6  \\
Fine-tuned (FLIR -- full shot) & RGB + IR & 26.5  & 48.4  \\
\bottomrule
\end{tabular}
}
\caption{Comparison of Grounding DINO open-set detection performance on M3FD between RGB zero-shot and FLIR fine-tuned models.}
\label{tab:open_set_m3fd}
\end{table}

\begin{table*}[t!]
\centering
\begin{minipage}[t]{0.45\linewidth}
\centering
\small
\scalebox{1}{
\begin{tabular}{@{}llc@{}}
\toprule
Model & Fusion & mAP50 \\
\midrule
\multirow{3}{*}{MS-YOLOW-M}
 & Sum                 & \textbf{71.15}\std{0.59} \\
 & Conv (random init)  & 56.14\std{3.07} \\
 & Conv (sum-init)     & \underline{68.69}\std{0.79} \\
\bottomrule
\end{tabular}
}
\caption{Impact of fusion strategy and initialization on FLIR (10-shot). mAP50 reported over 10 runs.}
\label{tab:flir_fusion_study_supp}
\end{minipage}
\hfill
\begin{minipage}[t]{0.4\linewidth}
\centering
\small
\setlength{\tabcolsep}{4pt}
\scalebox{1}{
\begin{tabular}{@{}lccc@{}}
\toprule
Method & Bicycle & Car & Person \\
\midrule
FSMODNet    & 27.05 & 32.04 & 30.67 \\
MS-YOLOW-M  & 58.00 & 79.01 & 75.05 \\
MS-GDINO-T  & 57.63 & 78.37 & 73.67 \\
\bottomrule
\end{tabular}
}
\caption{5-shot mAP50 on FLIR, RGB+IR, comparing FSMODNet with our methods.}
\label{tab:flir_fsmodnet_perclass}
\end{minipage}
\end{table*}

\subsection{Comparison with FSMODNet}
\label{sec:fsmodnet_comp}
\cref{tab:flir_fsmodnet_perclass} reports a direct comparison between 
our VLM-based framework and FSMODNet~\cite{nkegoum2025fsmodnet}. Our 
method outperforms FSMODNet across all evaluated settings, which is 
expected given the substantially richer semantic representations provided 
by large-scale VLM pretraining. We note an important methodological 
difference: FSMODNet follows the standard two-stage training protocol, 
where the model is first pre-trained on base classes before being 
fine-tuned on novel classes. In contrast, our VLM-based framework 
bypasses this two-stage procedure entirely, directly evaluating on 
few-shot novel classes without any base class pre-training phase, making 
our setting strictly harder in terms of task-specific supervision.

\subsection{Sum vs.\ $1{\times}1$ Convolution Fusion}
\label{sec:random}
To investigate the large performance gap between sum and $1{\times}1$ 
conv fusion (\cref{tab:flir_fusion_study}), we tested a sum-initialized 
$1{\times}1$ conv where weights are initialized as a per-channel 
identity map. Performance converges to sum's level, and learned weights 
remain within $10^{-4}$ of the identity initialization 
(diag.\ $\approx$0.9996--1.0001, 
off-diag.\ $\approx10^{-4}$--$10^{-6}$), confirming the model finds no 
benefit in deviating from sum-like fusion and that the gap is not due to 
optimization instability or initialization, but rather an inherent 
limitation of learned fusion under few-shot supervision.

\section{Visualization of Predictions}
\label{sec:visu}

\begin{figure*}[tb]
\centering
\includegraphics[width=\textwidth]{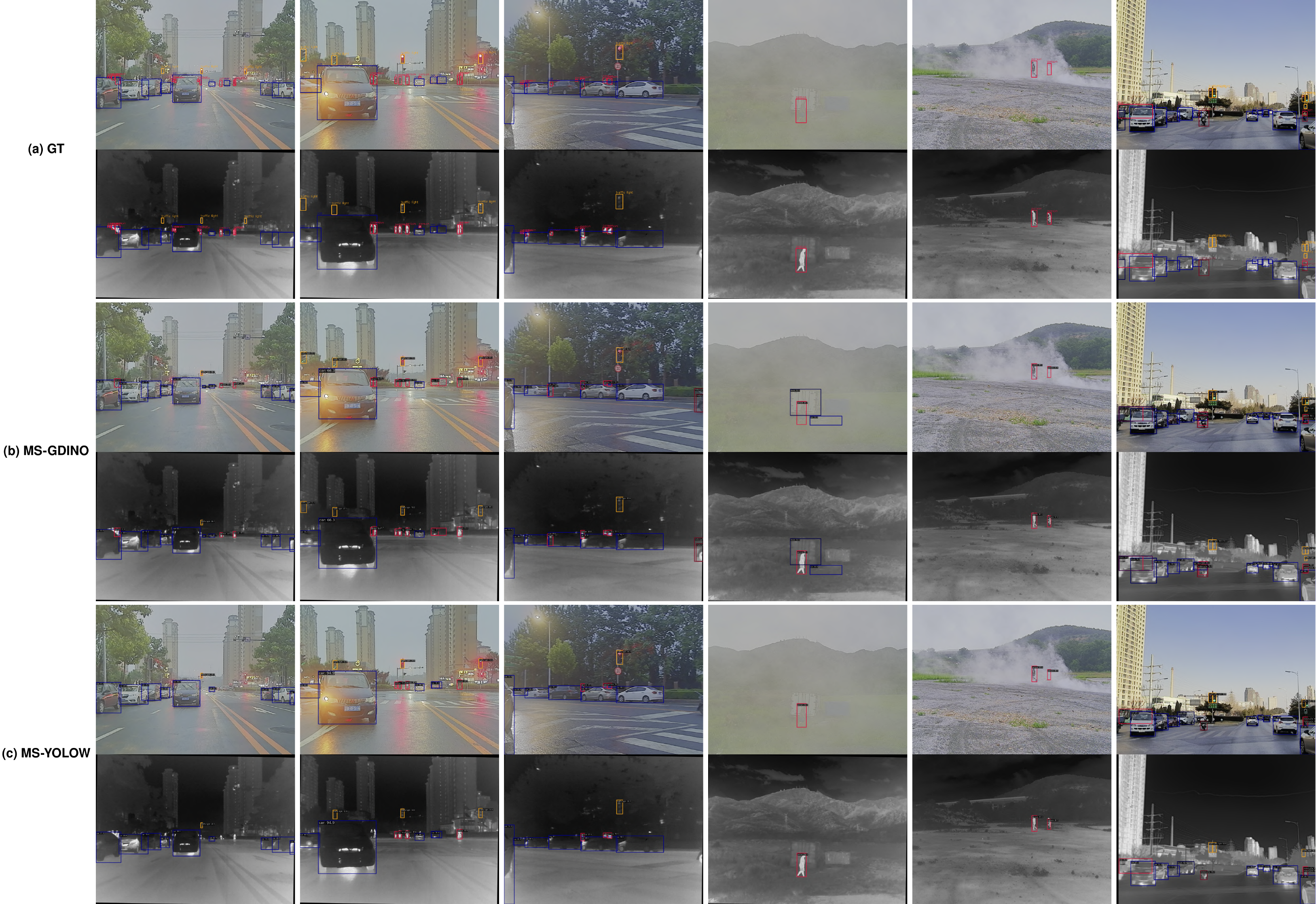}
\caption{Qualitative comparison of object detection results on challenging M3FD scenarios. The top part (a) shows the Ground Truth (GT) annotations on the RGB and IR images (first and second line, respectively). The middle part (b) presents the predictions from the MS-GDINO. The bottom part (c) displays the results from MS-YOLOW method. Models were fine-tuned with 5 shots. }
\label{fig:open_set_vis}
\end{figure*}

We visualize detection results on representative M3FD scenes in \cref{fig:open_set_vis}. 
MS-GDINO reliably detects most ground-truth objects across challenging conditions including low light, distant targets, and crowded scenes but occasionally produces false positives, such as misidentifying background structures (e.g., predicting a truck instead of a container in the third pair). 
In contrast, MS-YOLOW behaves more conservatively: it yields fewer false positives but misses several clear objects in crowded or long-range scenarios, as seen in the first pair, although it performs well in low-light scenes.  Overall, the qualitative comparison suggests that MS-GDINO favors higher recall at the cost of occasional false alarms, while MS-YOLOW prioritizes precision but struggles with crowded and distant targets.

\end{document}